\documentclass[final,5p,times,twocolumn]{elsarticle}

\usepackage{colortbl}
\usepackage{color}
\usepackage{array}
\usepackage{amsmath, amssymb}
\usepackage{nccmath}
\usepackage{booktabs}
\usepackage{pifont}
\newcommand{\cmark}{\ding{51}}%
\newcommand{\xmark}{\ding{55}}%
\usepackage{multirow}
\usepackage[lined,ruled,vlined, linesnumbered]{algorithm2e}
\SetKwComment{Comment}{$\triangleright$\ }{}
\newcommand{\update}{\textit{MemoryUpdate}}
\newcommand{\retrieval}{\textit{MemoryRetrieval}}
\newcommand{\modelupdate}{\textit{ModelUpdate}}
\usepackage{appendix}
\usepackage{chngcntr}   
\newcommand\Tstrut{\rule{0pt}{2.6ex}}         
\newcommand\Bstrut{\rule[-0.9ex]{0pt}{0pt}}   
\usepackage{subcaption}
\usepackage[utf8x]{inputenc} 
\usepackage{bbm}
\usepackage{commath}
\usepackage{tabularx}
\usepackage{algpseudocode}
\usepackage[table]{xcolor}
\usepackage{hyperref}

\def\boxit#1{%
  \smash{\fboxsep=1pt\llap{\rlap{\fbox{\strut\makebox[#1]{}}}~}}\ignorespaces
}

\journal{Neurocomputing}

\begin{document}
\captionsetup[figure]{labelfont={bf},name={Fig.},labelsep=period} 
\captionsetup[table]{labelfont={bf},name={Table},labelsep=period}
\begin{frontmatter}



\title{Online Continual Learning in Image Classification: \\An Empirical Survey}


\author[affiliation1,email1]{Zheda Mai\corref{mycorrespondingauthor}}
\cortext[mycorrespondingauthor]{Corresponding author}
\ead{zheda.mai@mail.utoronto.ca}
\author[affiliation1,email1]{Ruiwen Li}
\author[affiliation1,email2]{Jihwan Jeong}
\author[affiliation1,email1]{David Quispe}
\author[affiliation2,email3]{\\Hyunwoo Kim}
\author[affiliation1,email2]{Scott Sanner}

\address[affiliation1]{Department of Mechanical and Industrial Engineering, University of Toronto, Toronto, 5 King's College Road, ON M5S3G8, Canada}
\address[email1]{(zheda.mai, ruiwen.li, david.quispe)@mail.utoronto.ca}
\address[email2]{(jhjeong, ssanner)@mie.utoronto.ca}
\address[affiliation2]{LG AI Research, 128, Yeoui-daero, Yeongdeungpo-gu, Seoul, South Korea}
\address[email3]{hwkim@lgresearch.ai}

\begin{abstract}

Online continual learning for image classification studies the problem of learning to classify images from an online stream of data and tasks, where tasks may include new classes (class incremental) or data nonstationarity (domain incremental).  One of the key challenges of continual learning is to avoid catastrophic forgetting (CF), i.e., forgetting old tasks in the presence of more recent tasks. Over the past few years, a large range of methods and tricks have been introduced to address the continual learning problem, but many have not been fairly and systematically compared under a variety of realistic and practical settings.

To better understand the relative advantages of various approaches and the settings where they work best, this survey aims to (1) compare state-of-the-art methods such as Maximally Interfered Retrieval (MIR), iCARL, and GDumb (a very strong baseline) and determine which works best at different memory and data settings as well as better understand the key source of CF; (2) determine if the best online class incremental methods are also competitive in the domain incremental setting; and (3) evaluate the performance of 7 simple but effective tricks such as the ``review'' trick and the nearest class mean (NCM) classifier to assess their relative impact. Regarding (1), we observe that iCaRL remains competitive when the memory buffer is small; GDumb outperforms many recently proposed methods in medium-size datasets and MIR performs the best in larger-scale datasets. For (2), we note that GDumb performs quite poorly while MIR --- already competitive for (1) --- is also strongly competitive in this very different (but important) continual learning setting. Overall, this allows us to conclude that MIR is overall a strong and versatile online continual learning method across a wide variety of settings. Finally for (3), we find that all tricks are beneficial, and when augmented with the ``review'' trick and NCM classifier, MIR produces performance levels that bring online continual learning much closer to its ultimate goal of matching offline training. Our codes are available at \url{https://github.com/RaptorMai/online-continual-learning}.

\end{abstract}



\end{frontmatter}


\section{Introduction}
\label{intro}



With the ubiquity of personal smart devices and image-related applications, a massive amount of image data is generated daily. The need for updating the deep learning model with new data on personal (i.e., edge) devices to preserve privacy, minimize communication bandwidth and maintain real-time performance necessitates the development of methods that can continually learn from streaming data while minimizing memory storage and computation footprint.
However, incrementally updating a neural network with a nonstationary data stream results in \emph{catastrophic forgetting} (CF)~\cite{CF, goodfellow2013empirical} --- the inability of a network to perform well on previously seen data after updating with recent data. For this reason, conventional deep learning tends to focus on offline training, where each mini-batch is sampled i.i.d from a static dataset with multiple epochs over the training data. However, to accommodate changes in the data distribution, such a training scheme requires entirely retraining the network on the new dataset, which is inefficient and sometimes infeasible when previous data are not available due to storage limits or privacy issues. 


\textbf{\emph{Continual Learning}} (CL) studies the problem of learning from a non-i.i.d stream of data, with the goal of preserving and extending the acquired knowledge. A more complex and general viewpoint of CL is the stability-plasticity dilemma~\cite{dilemma1, dilemma2}, where stability refers to the ability to preserve past knowledge and plasticity denotes the fast adaptation of new knowledge. Following this viewpoint, CL seeks to strike a balance between learning stability and plasticity. Since CL is often used interchangeably with lifelong learning~\cite{chen2018lifelong, yoon2018lifelong} and incremental learning~\cite{rebuffi2017icarl, chaudhry2018riemannian}, for simplicity, we will use CL to refer to all concepts mentioned above. 

Conventional CL assumes that new data arrive one task at a time, and the data distribution for each task is stationary~\cite{survey}. Thus, a model can be trained in an \emph{offline} manner, namely multiple epochs over the current task with repeat shuffle. However, this setting requires storing all data from the current task for training, which may not be feasible due to privacy issues or resource limitations. 

In this survey, we focus on a more realistic but challenging problem, \textbf{\emph{Online Continual Learning}}, where data arrive one tiny batch at a time and previously seen batches from the current or the previous tasks are not accessible. Therefore, a model is required to efficiently learn from a single pass over the online data stream where the model may experience new classes (\emph{Online Class Incremental, \textbf{OCI}}) or data nonstationarity, including new background, blur, noise, illumination, and occlusion, etc.(\emph{Online Domain Incremental, \textbf{ODI}}). Moreover, there are two types of inference configuration in CL, multi-head, and single-head~\cite{chaudhry2018riemannian}. With the multi-head configuration, a dedicated head (classification layer) is assigned to each task, and the model just needs to classify labels within a task. Although multi-head configuration is adopted to online continual learning in some works~\cite{agem, gem, chen2020mitigating, pham2020contextual}, the requirement of additional supervisory signals at inference --- namely the task-ID --- to select the corresponding head obviates its use when the task-ID is unavailable. Thus, in this work, we adopt the single-head configuration where the model has a shared output layer, and it needs to classify all labels without task-IDs. The axes of our setting are based on the practical CL desiderata proposed recently~\cite{robot, survey, robust} and have received much attention in the past years~\cite{mir, gss, dirichlet}.

To keep this paper focused, we only consider the supervised classification problem in computer vision. Although CL is also studied in reinforcement learning~\cite{er_rl, rl_meta, rl_cl_synapses} and more recently in unsupervised learning~\cite{unsuper}, the image classification problem is still the main focus for many CL researchers.


Over the past few years, a broad range of methods and tricks have been introduced to address CL problems, but many have not been fairly and systematically compared under a variety of settings. To better understand the relative advantages of different approaches and the settings where they work best, this survey aims to do the following:
\begin{itemize}
    \item We fairly compare state-of-the-art methods in OCI and determine which works best at different memory and data settings. We observe that iCaRL~\cite{rebuffi2017icarl} (proposed in 2017) remains competitive when the memory buffer is small; GDumb~\cite{prabhu2020gdumb} is a strong baseline that outperforms many recently proposed methods in medium-size datasets, while MIR~\cite{mir} performs the best in a larger-scale dataset. Also, we experimentally and theoretically confirm that a key cause of CF is due to the recency learning bias towards new classes in the last fully connected layer owing to the imbalance between previous data and new data.
    \item We determine if the best OCI methods are also competitive in the ODI setting. We note that GDumb performs quite poorly in ODI, whereas MIR --- already competitive in OCI --- is still strongly competitive in ODI. Overall, these results allow us to conclude that MIR is a strong and versatile online CL method across a wide variety of settings.
    \item We evaluate the performance of 7 simple but effective tricks to assess their relative impacts.  We find that all tricks are beneficial and when augmented with the ``review'' trick~\cite{zheda} and a nearest class mean (NCM) classifier~\cite{rebuffi2017icarl}, MIR produces performance levels that bring online CL much closer to its ultimate goal of matching offline training.
\end{itemize}


The remainder of this paper is organized as follows. Section~\ref{related} discusses the existing surveys in the CL community, and Section~\ref{definition} formally defines the problem, settings and evaluation metrics. In Section~\ref{hyperparameter}, we explain the online continual hyperparameter tuning method we use. Then, Section~\ref{approaches} provides an overview of state-of-the-art CL techniques, while Section~\ref{methods} gives a detailed description of methods that we compared in experiments. We discuss how class imbalance results in catastrophic forgetting and introduce CL tricks that can effectively alleviate the forgetting in Section~\ref{trick}. We outline our experimental setup, comparative evaluation and key findings in Section~\ref{experiments}. Finally, Section~\ref{future} discusses recent trends and emerging directions in CL, and we conclude in Section~\ref{conclusion}.



\section{Related Work}
\label{related}
With the surge in the popularity of CL, there are multiple reviews and surveys covering the advances of CL. The first group of surveys are not empirical.  \cite{review} discusses the biological perspective of CL and summarizes how various approaches alleviate catastrophic forgetting. \cite{robot} formalizes the CL problem and outlines the existing benchmarks, metrics, approaches and evaluation methods with the emphasis on robotics applications. They also recommend some desiderata and guidelines for future CL research. \cite{parisi2020online} emphasizes the importance of online CL and discusses recent advances in this setting. Although these three surveys descriptively review the recent development of CL and provide practical guidelines, they do not perform any empirical comparison between methods. 

In contrast, the second group of surveys on CL are empirical. For example, \cite{reevaluate, three} evaluate multiple CL methods on three CL scenarios: task incremental, class incremental and domain incremental. \cite{robust} empirically analyzes and criticizes some common experimental settings, including the multi-head configuration~\cite{chaudhry2018riemannian} with an exclusive output layer for each task and the use of permuted-type datasets (e.g., permuted MNIST). However, the analysis in these three works is limited to small datasets such as MNIST and Fashion-MNIST. Another two empirical studies on the performance of CL include \cite{comprehensive, MeasuringCF},  but only a small number of CL methods are compared. The first extensive comparative CL survey with empirical analysis is presented in \cite{survey}, which focuses on the task incremental setting with the multi-head configuration. In contrast, our work addresses a more practical and realistic setting, Online Continual Learning with single-head configuration, which require the model to learn online without access to task-ID at training and inference time. Also, we summarize and evaluate several simple but effective tricks to assess their relative impacts. 
%
%
%

\begin{table*}
\scriptsize
\centering
\begin{tabular}{c | c c c | c|c} 
    \toprule
    \multirow{2}{*}{Scenario}&\multicolumn{3}{c|}{Difference between $D_{i-1}$ and $D_{i}$}&\multirow{2}{*}{Task-ID}&\multirow{2}{*}{Online}\\
     & $P(X_{i-1})\ne P(X_{t})$  &$P(Y_{i-1})\ne P(Y_{i})$& $\{Y_{i-1}\}\ne \{Y_{i}\}$ &\\ 
    \hline\hline

    Task Incremental& \cmark &\cmark&\cmark& Train \& Test&No\\
    Class Incremental&\cmark &\cmark&&No&Optional\\
    Domain Incremental &\cmark&&&No&Optional\\
    \bottomrule
\end{tabular}
\caption{Three continual learning scenarios  based on the difference between $D_{i-1}$ and $D_{i}$, following~\cite{reevaluate}. $P(X)$ is the input data distribution; $P(Y)$ is the target label distribution; $\{Y_{i-1}\}\ne \{Y_{i}\}$ denotes that output
space are from a disjoint space which is separated by task-ID.}
\label{tab:scenario}
\end{table*}
\section{Problem Definition and Evaluation Metrics}
\label{definition}
\subsection{Problem definition for Online Continual Learning}
\label{problem}
We consider the supervised image classification problem with an online (potentially infinite) non-i.i.d stream of data, following the recent CL literature \cite{mir,gss, parisi2020online, robot}. Formally, we define a data stream of unknown distributions $\mathcal{D} = \{D_1, \ldots, D_N\}$ over $X\times Y$, where $X$ and $Y$ are input and output random variables respectively, and a neural network classifier parameterized by $\theta$, $f:X \mapsto\mathbb{R}^C$ where $C$ is the number of classes observed so far as in~\cite{robot}. At time $t$, a CL algorithm $A^{CL}$ receives a mini-batch of samples ($\mathit{x_t^i}$, $\mathit{y_t^i}$) from the current distribution $D_i$, and the algorithm only sees this mini-batch once.

An algorithm $A^{CL}$ is defined with the following signature:
\begin{ceqn}
\begin{align}
A_{t}^{CL}:\ \langle f_{t-1}, (\mathit{x_t}, \mathit{y_t}), M_{t-1}\rangle \ \rightarrow \ \langle f_{t}, M_{t}\rangle
\end{align}
\end{ceqn}

Where:
\begin{itemize}
    \item $f_t$ is the classifier at time step $t$.
    \item ($\mathit{x_t^i}$, $\mathit{y_t^i}$) is a mini-batch received at time $t$ from $D_i$ which contains $\{(\mathit{x_{tj}^i}, \mathit{y_{tj}^i})\mid j \in[1, \ldots, b]\}$ where b is the mini-batch size.
    \item $M_t$ is an external memory that can be used to store a subset of the training samples or other useful data (e.g., the classifier from the previous time step as in LwF~\cite{lwf}). Note that the online setting does not limit the usage of the samples in $M$, and therefore, the classifier $f_t$ can use them as many times as it wants.  
\end{itemize}


Note that we assume, for simplicity, a locally i.i.d stream of data where each task distribution $D_i$ is stationary as in~\cite{gem, agem}; however, this framework can also accommodate the setting in which samples are drawn non-i.i.d from $D_i$ as in~\cite{gepperth2016incremental, hayes2018new}, where concept drift may occur within $D_i$.

The goal of $A^{CL}$ is to train the classifier $f$ to continually learn new samples from the data stream without interfering with the performance of previously observed samples. Note that unless the current samples are stored in $M_t$, $A^{CL}$ will not have access to these sample in the future. Formally, at time step $\tau$, $A^{CL}$ tries to minimize the loss incurred by all the previously seen samples with only access to the current mini-batch and data from $M_{\tau-1}$:
\begin{ceqn}
\begin{align}
\mathrm{min}_\theta \sum_{t=1}^{\tau} \mathbb{E}_{\left(\mathit{x}_{t}, \mathit{y}_{t}\right)}\left[\ell\left(f_{\tau}\left(\mathit{x}_{t} ; \theta\right), \mathit{y}_{t}\right)\right]
\end{align}
\end{ceqn}

\begin{table*}[h!]
\footnotesize
\begin{center}
\begin{tabular}{|c|m{0mm}cc| m{0mm}cc|m{0mm}cc|}
\hline
Task&\multicolumn{3}{c|}{Task Incremental} &\multicolumn{3}{c|}{Class Incremental}&\multicolumn{3}{c|}{Domain Incremental} \\ \hline
\multirow{2}{*}{$D_{i-1}$}
&
x:&
\raisebox{-.4\height}{\includegraphics[width=0.09\textwidth ]{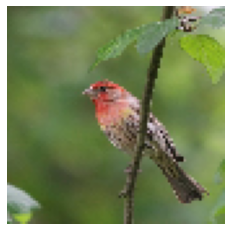}}&
\raisebox{-.4\height}{\includegraphics[width=0.09\textwidth]{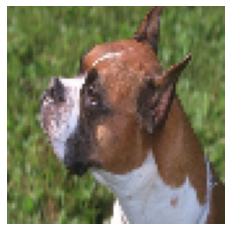}}&
x:&
\raisebox{-.4\height}{\includegraphics[width=0.09\textwidth ]{pics/bird.png}}&
\raisebox{-.4\height}{\includegraphics[width=0.09\textwidth]{pics/dog.png}}&
x:&
\raisebox{-.4\height}{\includegraphics[width=0.09\textwidth ]{pics/bird.png}}&
\raisebox{-.4\height}{\includegraphics[width=0.09\textwidth]{pics/dog.png}}
\\
&y:& Bird & Dog
&y:& Bird & Dog
&y:& Bird & Dog
\\
\cline{1-10}
task-ID(test)&\multicolumn{3}{c|}{\textbf{i-1}}
&\multicolumn{3}{c|}{\textbf{Unknown}}
&\multicolumn{3}{c|}{\textbf{Unknown}}
\\
\hline
\multirow{3}{*}{$D_{i}$}
&
x:&
\raisebox{-.4\height}{\includegraphics[width=0.09\textwidth]{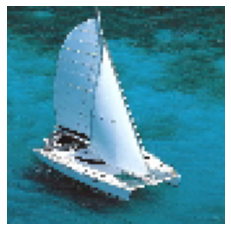}}&
\raisebox{-.4\height}{\includegraphics[width=0.09\textwidth]{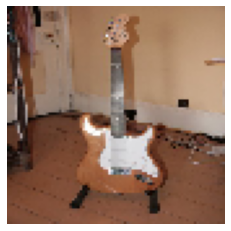}}&
x:&
\raisebox{-.4\height}{\includegraphics[width=0.09\textwidth]{pics/ship.png}}&
\raisebox{-.4\height}{\includegraphics[width=0.09\textwidth]{pics/guita.png}}&
x:&
\raisebox{-.4\height}{\includegraphics[width=0.09\textwidth]{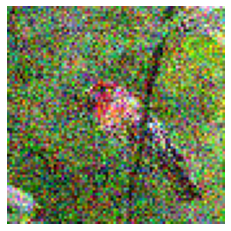}}&
\raisebox{-.4\height}{\includegraphics[width=0.09\textwidth]{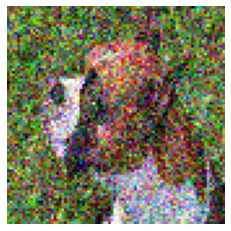}}
\\
&y:& Ship & Guitar
&y:& Ship & Guitar
&y:& Bird & Dog
\\
\cline{1-10}
task-ID(test)&\multicolumn{3}{c|}{\textbf{i}}
&\multicolumn{3}{c|}{\textbf{Unknown}}
&\multicolumn{3}{c|}{\textbf{Unknown}}
\\\hline
\end{tabular}

\caption{Examples of the three CL scenarios. (x, y, task-ID) represents (input images, target label and task identity). The main distinction between task incremental and class incremental is the availability of task-ID. The main difference between class incremental and domain incremental is that, in class incremental, a new task contains completely new classes, whereas domain incremental, a new task consists of new instances with nonstationarity (e.g., noise) of all the seen classes.}
\label{tab: scenario-example}
\end{center}
\end{table*}

Recently, \cite{reevaluate, three} have categorized the CL problem into three scenarios based on the difference between $D_{i-1}$ and $D_{i}$. Table~\ref{tab:scenario} summarizes the differences between the three scenarios, i.e., task incremental, class incremental and domain incremental. For task incremental, the output spaces are separated by task-IDs and are disjoint between $D_{i-1}$ and $D_{i}$. We denote this setting as $\{Y_{i-1}\}\ne \{Y_{i}\}$, which in turn leads to $P(Y_{i-1})\ne P(Y_{i})$. In this setting, task-IDs are available during both train and test times. For class incremental, mutually exclusive sets of classes comprise each data distribution $D_i$, meaning that there is no duplicated class among different task distributions.
Thus $P(Y_{i-1})\ne P(Y_{i})$, but the output space is the same for all distributions since this setting adopts the single-head configuration where the model needs to classify all labels without a task-ID. Domain incremental represents the setting where input distributions are different, while the output spaces and distribution are the same. Note that task IDs are not available for both class and domain incremental. Table~\ref{tab: scenario-example} shows examples of these three scenarios. Following this categorization, the settings we focus on in this work are known as Online Class Incremental (OCI) and Online Domain Incremental (ODI) with the single-head configuration. Note that some works consider the relaxed version of OCI with the multi-head configuration~\cite{gem, agem, pham2020contextual}, especially for methods without using memory buffer~\cite{fini2020online, chen2020mitigating, pham2020contextual}. 

\subsection{Evaluation metrics}
\label{metrics}

Besides measuring the final accuracy across tasks, it is also critical to assess how fast a model learns, how much the model forgets and how well the model transfers knowledge from one task to another. To this end, we use five standard metrics in the CL literature to measure performance: (1) the average accuracy for overall performance~\cite{agem}; (2) the average forgetting to measure how much of the acquired knowledge the model has forgotten~\cite{chaudhry2018riemannian}; (3) the forward transfer and (4) the backward transfer to assess the ability for knowledge transfer~\cite{robot, gem}; (5) the total running time, including training and testing times.

Formally, we define $a_{i,j}$ as the accuracy evaluated on the held-out test set of task $j$ after training the network from task 1 through to $i$, and we assume there are $T$ tasks in total.

\textbf{Average Accuracy} can be defined as Eq.~\eqref{eq: acc}. When $i=T$, $A_T$ represents the average accuracy by the end of training with the whole data sequence (see example in Table~\ref{tab: metric}).
\begin{ceqn}
\begin{align}
\text{Average Accuracy} (A_{i})=\frac{1}{i} \sum_{j=1}^{i} a_{i, j}
\label{eq: acc}
\end{align}
\end{ceqn}

\textbf{Average Forgetting} at task $i$ is defined as Eq.~\eqref{eq: fgt}. $f_{i, j}$ represents how much the model has forgot about task $j$ after being trained on task $i$. Specifically, $\max\limits_{l \in\{1, \cdots, k-1\}}(a_{l, j})$ denotes the best test accuracy the model has ever achieved on task j before learning task $k$, and $a_{k, j}$ is the test accuracy on task $j$ after learning task $k$. 

\begin{ceqn}
\begin{gather} 
\text{Average Forgetting} (F_{i})=\frac{1}{i-1} \sum_{j=1}^{i-1} f_{i, j} \nonumber\\
\text{where~} f_{k, j}=\max_{l \in\{1, \cdots, k-1\}}(a_{l, j})-a_{k, j}, \forall j<k
\label{eq: fgt}
\end{gather}
\end{ceqn}

\begin{table}[]
\centering
\begin{tabular}{l|ccccc}
\toprule
a & $te_1$ &$te_2$  &$\dots$ &$te_{T-1}$&$te_T$  \\ \midrule
$tr_1$ & $a_{1,1}$ & \cellcolor{green!25}$a_{1,2}$&  \cellcolor{green!25}$\dots$&\cellcolor{green!25} $a_{1,T-1}$&\cellcolor{green!25} $a_{1,T}$ \\
$tr_2$ & \cellcolor{blue!25}$a_{2,1}$ & $a_{2,2}$& \cellcolor{green!25}$\dots$& \cellcolor{green!25}$a_{2,T-1}$& \cellcolor{green!25}$a_{2,T}$\\

$\dots$ & \cellcolor{blue!25}$\dots$ &\cellcolor{blue!25}$\dots$ &$\dots$ &\cellcolor{green!25}$\dots$&\cellcolor{green!25}$\dots$\\

$tr_{T-1}$ & \cellcolor{blue!25}$a_{T-1,1}$ & \cellcolor{blue!25} $a_{T-1,2}$& \cellcolor{blue!25}$\dots$&$a_{T-1,T-1}$&\cellcolor{green!25}$a_{T-1,T}$\\

$tr_T$ & \boxit{2.25in}\cellcolor{blue!25}$a_{T,1}$ & \cellcolor{blue!25} $a_{T,2}$& \cellcolor{blue!25}$\dots$&\cellcolor{blue!25}$a_{T,T-1}$&$a_{T,T}$\\ \bottomrule
\end{tabular}
\caption{Accuracy matrix example following the notations in~\cite{robot}. $tr_i$ and $te_i$ denote training and test set of task $i$. $A_T$ is the average of accuracies in the box. $BWT^+$ is the average of accuracies in purple and $FWT$ is the average of accuracies in green.  }
\label{tab: metric}
\end{table}

\textbf{Positive Backward Transfer}($BWT^+$) measures the positive influence of learning a new task on preceding tasks' performance (see example in Table~\ref{tab: metric}). 
\begin{ceqn}
\begin{align}
BWT^+=max(\frac{\sum_{i=2}^{T} \sum_{j=1}^{i-1}\left(a_{i, j}-a_{j, j}\right)}{\frac{T(T-1)}{2}}, 0)
\label{eq: bwt}
\end{align}
\end{ceqn}

\textbf{Forward Transfer}($FWT$) measures the positive influence of learning a task on future tasks' performance (see example in Table~\ref{tab: metric}). 
\begin{ceqn}
\begin{align}
F W T=\frac{\sum_{i=1}^{j-1} \sum_{j=1}^{T} a_{i, j}}{\frac{T(T-1)}{2}}
\label{eq: fwt}
\end{align}
\end{ceqn}



\section{Overview of Continual Learning Techniques}
\label{approaches}
Although a broad range of methods have been introduced in the past few years to address the CL problem, the assumptions that each method makes are not consistent due to the plethora of settings in CL.
In particular, some methods have a better ability to generalize to different CL settings because they require less supervisory signals during both training and inference times. For example, ER~\cite{tiny} was designed for the task incremental setting, but the method can easily be used in all the other CL settings since it does not need any additional supervisory signals. In this sense, a systematic summary of supervisory signals that different methods demand will help understand the generalization capacity and limitations of the methods. Furthermore, the summary will facilitate fair comparison in the literature. On the other hand, CL methods have typically been taxonomized into three major categories based on the techniques they use: regularization-based, memory-based and parameter-isolation-based~\cite{survey, review}. A clear trend in recent works, however, is to simultaneously apply multiple techniques in order to tackle the CL problem. In this section, we comprehensively summarize recently proposed methods based on techniques they use and supervisory signals required at training and inference times (see Table~\ref{tab: summary}). 

\paragraph{Supervisory Signals} The most important supervisory signal is the task-ID. When a task-ID is available, a training or testing sample is given as $(x, y, t)$ instead of $(x, y)$ where $t$ is the task-ID. For the task incremental setting, task-IDs are available at both training and inference times. In regards to the class incremental setting, a task-ID is not available at the inference time but can be inferred during training as each task has disjoint class labels. On the other hand, in the domain incremental setting, a task-ID is not available at all times. Other supervisory signals include a natural language description of a task or a matrix specifying the attribute values of the objects to be recognized in the task~\cite{agem}. 



Moreover, a method is \textit{online-able} if it does not need to revisit samples it has processed before. Hence, to be online-able, the model needs to learn efficiently from one pass of the data. For example, the herding-based memory update strategy proposed in iCaRL~\cite{rebuffi2017icarl} needs all the samples from a class to select a representative set; therefore, methods using this strategy are not online-able.

\paragraph{Techniques} 
\textbf{Regularization} techniques impose constraints on the update of network parameters to mitigate catastrophic forgetting. This is done by either incorporating additional penalty terms into the loss function~\cite{imm, mas, si, laplace} or modifying the gradient of parameters during optimization~\cite{gem, agem, he2018overcoming}. 

\textbf{Knowledge Distillation (KD)}~\cite{distill} is an effective way for knowledge transfer between networks. KD has been widely adopted in CL methods~\cite{EEIL, lwf, wu2019large, encoder}, and it is often considered as one of the regularization techniques. Due to the prevalence of KD in CL methods, we list it as a separate technique in Table~\ref{tab: summary}. One shortcoming of regularization-based techniques including KD is that it is difficult to strike a balance between the regularization and the current learning when learning from a long stream of data.

\textbf{Memory}-based techniques store a subset of samples from previous tasks for either replay while training on new task~\cite{tiny, mir, gss} or regularization purpose~\cite{vcl, gem, tao2020topology}. These methods become infeasible when storing raw samples is not possible due to privacy or storage concerns. 

Instead of saving raw samples, an alternative is \textbf{Generative Replay} which trains a deep generative model such as GAN~\cite{gan} to generate pseudo-data that mimic past data for replay~\cite{generative, gen_incre, van2018generative}. The main disadvantages of generative replay are that it takes long time to train such generative models and that it is not a viable option for more complex datasets given the current state of deep generative models~\cite{lesort2019generative, mir}. 

\textbf{Parameter-isolation (PI)}-based techniques bypass interference by allocating different parameters to each task. PI can be subdivided into Fixed Architecture (FA) that only activates relevant parameters for each task without modifying the architecture~\cite{mallya2018packnet, fernando2017pathnet, hat}, and Dynamic Architecture (DA) that adds new parameters for new tasks while keeping old parameters unchanged~\cite{rusu2016progressive, den, aljundi2017expert}. Most previous works require task-IDs at inference time, and a few recent methods have been introduced to predict without a task-ID~\cite{dirichlet, random_path}. 

For a more detailed discussion of these techniques, we refer readers to the recent CL surveys~\cite{survey, review, robot}.

\begin{table*}[t!]
\centering
\scriptsize
\setlength{\tabcolsep}{0pt} 
\renewcommand{\arraystretch}{1.7}
\begin{tabular*}{\textwidth}{@{\extracolsep{\fill}\quad}lccccccccc}
\toprule

 & \multicolumn{3}{c}{Settings} & \multicolumn{6}{c}{Techniques}\\
\cmidrule{2-4}\cmidrule{5-10}

Methods & t-ID free(test) & t-ID free(train) & Online-able & Reg & Mem & KD & PI(FA) & PI(DA) & Generative\\
\midrule    

MIR\cite{mir}, GSS\cite{gss}, ER\cite{tiny}, CBO\cite{borsos2020coresets} & \multirow{13}{*}{\cmark} & \multirow{13}{*}{\cmark} & \multirow{13}{*}{\cmark} & \multirow{4}{*}{} & \multirow{4}{*}{\cmark} & \multirow{4}{*}{} & \multirow{4}{*}{} & \multirow{4}{*}{} & \multirow{4}{*}{} \\
GDumb\cite{prabhu2020gdumb}, DER\cite{buzzega2020dark}, MER\cite{mer}\\
CBRS\cite{cbrs}, GMED\cite{gmed}, PRS\cite{prs}\\
La-MAML\cite{gupta2020maml}, MEFA\cite{iscen2020memory}\\
\cline{5-10}
MERLIN\cite{merlin} & & & &&\cmark & &\cmark&&\\
\cline{5-10}
CN-DPM\cite{dirichlet}, TreeCNN\cite{roy2020tree} & & & & & & &&  \cmark &\\
\cline{5-10}
A-GEM\cite{agem}, GEM\cite{gem}, VCL~\cite{vcl} & & & & \cmark & \cmark & & & & \\
\cline{5-10}
WA\cite{wa}, BiC\cite{wu2019large}, LUCIR\cite{unified}& & & && \multirow{2}{*}\cmark &\multirow{2}{*} \cmark  & & &\\
IL2M\cite{il2m}, ILO\cite{he2020incremental}\\
\cline{5-10}
LwF-MC\cite{rebuffi2017icarl}, LwM\cite{dhar2019learning}, DMC\cite{dmc}& & & &&& \cmark  & & &\\
\cline{5-10}
SRM\cite{riemer2019scalable}, AQM\cite{caccia2019online}& & & & &\cmark && & &\cmark\\
\cline{5-10}
EWC++\cite{chaudhry2018riemannian} & & && \cmark & & & & &\\
\cline{5-10}
AR1\cite{maltoni2019continuous} & & && \cmark & & &\cmark & &\\


\hline

EEIL\cite{EEIL}, iCaRL\cite{rebuffi2017icarl}, MCIL\cite{liu2020mnemonics}& \multirow{4}{*}{\cmark} & \multirow{4}{*}{\cmark} & \multirow{4}{*}{\xmark} && \cmark & \cmark  & & &\\
\cline{5-10}
SDC\cite{yu2020semantic}& & & & \cmark & & & & &\\
\cline{5-10}
DGR\cite{generative}& & & && & & & & \cmark \\
\cline{5-10}
DGM\cite{dgm}& & & && & & \cmark&\cmark&\cmark\\
\cline{5-10}
ICGAN\cite{wu2018incremental_gan}, RtF\cite{van2018generative}& & & && &\cmark & & &\cmark\\
\hline
iTAML\cite{rajasegaran2020itaml},CCG\cite{abati2020conditional} & \cmark& \xmark &\cmark&  & \cmark&& \cmark&\\
\hline
\Tstrut

\end{tabular*}
\caption{Summary of recently proposed CL methods based on supervisory signals required and techniques they use. t-ID free(test/train) means task-ID is not required at test/train time.}
\label{tab: summary}
\end{table*}

\section{Compared Methods}
\label{methods}
\subsection{Regularization-based methods}
\paragraph{\bf Elastic Weight Consolidation (EWC)}EWC~\cite{Kirkpatrick2017} incorporates a quadratic penalty to regularize the update of model parameters that were important to past tasks. The importance of parameters is approximated by the diagonal of the Fisher Information Matrix $F$. Assuming a model sees two tasks A and B in sequence, the loss function of EWC is:
\begin{ceqn}
\begin{align}
\mathcal{L}(\theta)=\mathcal{L}_{B}(\theta)+\sum_{j} \frac{\lambda}{2} F_{j}\left(\theta_{j}-\theta_{A, j}^{*}\right)^{2}
\label{eq:ewc}
\end{align}
\end{ceqn}
where $\mathcal{L}_{B}(\theta)$ is the loss for task B, $\theta^*_{A, j}$ is the optimal value of $j^{th}$ parameter after learning task A and $\lambda$ controls the regularization strength. There are three major limitations of EWC: (1) It requires storing the Fisher Information Matrix for each task, which makes it impractical for a long sequence of tasks or models with millions of parameters. (2) It needs an extra pass over each task at the end of training, leading to its infeasibility for the online CL setting. (3) Assuming the Fisher to be diagonal may not be accurate enough in practice. Several variants of EWC are proposed lately to address these limitations~\cite{schwarz2018progress, chaudhry2018riemannian, rotateEWC}. As we use the online CL setting, we compare EWC++, an efficient and online version of EWC that keeps a single Fisher Information Matrix calculated by moving average. Specifically, given $F^{t-1}$ at $t-1$, the Fisher Information Matrix at $t$ is updated as:
\begin{ceqn}
\begin{align}
F^{t} = \alpha F_{tmp}^{t}+(1-\alpha) F^{t-1} 
\label{eq:ewc_online}
\end{align}
\end{ceqn}
where $F_{tmp}^{t}$ is the Fisher Information Matrix calculated with the current batch of data and $\alpha \in [0, 1]$ is a hyperparameter controlling the strength of favouring the current $F^t$.

\paragraph{\bf Learning without Forgetting (LwF)}LwF~\cite{lwf} utilizes knowledge distillation~\cite{distill} to preserve knowledge from past tasks in the multi-head setting~\cite{chaudhry2018riemannian}. In LwF, the teacher model is the model after learning the last task, and the student model is the model trained with the current task. Concretely, when the model receives a new task ($X_n$, $Y_n$), LwF computes $Y_o$, the output of old tasks for the new data $X_n$. During training, LwF optimizes the following loss:
\begin{equation}
\mathcal{L}(\theta) = \left(\lambda_{o} \mathcal{L}_{\text {KD}}\left(Y_{o}, \hat{Y}_{o}\right)+\mathcal{L}_{\text {CE}}\left(Y_{n}, \hat{Y}_{n}\right)+\mathcal{R}\left( \theta\right)\right)
\end{equation}
where $\hat{Y}_{o}$ and $\hat{Y}_{n}$ are the predicted values of the old task and new task using the same $X_n$. $\mathcal{L}_{\text {KD}}$ is the knowledge distillation loss incorporated to impose output stability of old tasks with new data and $\mathcal{L}_{\text {CE}}$ is the cross-entropy loss for the new task. $\mathcal{R}$ is a regularization term, and $\lambda_{o}$ is a hyperparameter controlling the strength of favouring the old tasks over the new task. A known shortcoming of LwF is its heavy reliance on the relatedness between the new and old tasks. Thus, LwF may not perform well when the distributions of the new and old tasks are different~\cite{lwf, rebuffi2017icarl, aljundi2017expert}. To apply LwF in the single-head setting where all tasks share the same output head, a variant of LwF (LwF.MC) is proposed in~\cite{rebuffi2017icarl}, and we evaluate LwF.MC in this work.

\subsection{Memory-based methods}
A generic online memory-based method is presented in Algorithm~\ref{alg:algo_er}. For every incoming mini-batch, the algorithm retrieves another mini-batch from a memory buffer, updates the model using both the incoming and memory mini-batches and then updates the memory buffer with the incoming mini-batch. What differentiate various memory-based methods are the memory retrieval strategy (line 3)~\cite{mir, aser}, model update (line 4)~\cite{agem, gem} and memory update strategy (line 5)~\cite{cbrs, prs, gss}. 

\begin{algorithm}
\footnotesize
\caption{Generic online Memory-based method}
\label{alg:algo_er}

\SetCustomAlgoRuledWidth{0.49\textwidth}
\SetAlgorithmName{Algorithm}{}{}
    \SetKwInOut{Input}{Input~}
    \SetKwInOut{Output}{Output}
    \SetKwInput{Initialize}{Initialize}
\Input{Batch size $b$, Learning rate $\alpha$}
\Initialize{Memory $\mathcal{M}\leftarrow \{\}*M$; Parameters $\theta$; Counter $n\leftarrow 0$ }
    \For{$t\in\{1,\dots,T\}$}{
        \For{$B_n\sim D_t$}{
        $B_\mathcal{M}\!\!\leftarrow\!$ \retrieval{}($B_n,\! \mathcal{M}$)\\
        $\theta\leftarrow~\text{\textit{ModelUpdate}}(B_n\cup B_\mathcal{M},\theta, \alpha)$ \\
        $\mathcal{M}\leftarrow$ \update{}$(B_n, \mathcal{M})$\\
        $n\leftarrow n + b$
        }
    }
     \textbf{return} $\theta$
\end{algorithm}

\paragraph{\bf Averaged GEM (A-GEM)} A-GEM~\cite{agem} is a more efficient version of GEM~\cite{gem}. Both methods prevent forgetting by constraining the parameter update with the samples in the memory buffer. At every training step, GEM ensures that the loss of the memory samples for each individual preceding task does not increase, while A-GEM ensures that the average loss for all past tasks does not increase. Specifically, let $g$ be the gradient computed with the incoming mini-batch and $g_{ref}$ be the gradient computed with the same size mini-batch randomly selected from the memory buffer. In A-GEM, if $g^Tg_{ref} \geq 0$, $g$ is used for gradient update but when $g^Tg_{ref} < 0$, $g$ is projected such that $g^Tg_{ref} = 0$. The gradient after projection is:
\begin{ceqn}
\begin{align}
\tilde{g}=g-\frac{g^{\top} g_{r e f}}{g_{r e f}^{\top} g_{r e f}} g_{r e f}
\end{align}
\end{ceqn}
As we can see, A-GEM focuses on \modelupdate{} in Algorithm~\ref{alg:algo_er}, and we apply reservoir sampling~\cite{reservoir} in \update{} and random sampling in \retrieval{} for A-GEM.

\paragraph{\bf Incremental Classifier and Representation Learning (iCaRL)} iCaRL~\cite{rebuffi2017icarl} is a replay-based method that decouples the representation learning and classification. For representation learning, the training set is constructed by mixing all the samples in the memory buffer and the current task samples. The loss function includes a classification loss to encourage the model to predict the correct labels for new classes and a KD loss to prompt the model to reproduce the outputs from the previous model for old classes. Note that the training set is imbalanced since the number of new-class samples in the current task is larger than that of the old-class samples in the memory buffer. iCaRL applies the binary cross entropy (BCE) for each class to handle the imbalance, but BCE may not be effective in addressing the relationship between classes. For the classifier, iCaRL uses a nearest-class-mean classifier~\cite{ncm} with the memory buffer to predict labels for test images. Moreover, it proposes a \update{} method based on the distance in the latent feature space with the inspiration from~\cite{welling2009herding}. For each class, it looks for a subset of samples whose mean of latent features are closest (in the Euclidean distance) to the mean of all the samples in this class. However, this method requires all samples from every class, and therefore it cannot be applied in the online setting. As such, we modify iCaRL to use reservoir sampling~\cite{reservoir}, which has been shown effective for \update{}~\cite{tiny}.

\paragraph{\bf Experience Replay (ER)} ER refers to a simple but effective replay-based method that has been discussed in~\cite{tiny, efficientER}. It applies reservoir sampling~\cite{reservoir} in \update{} and 
random sampling in \retrieval{}. Reservoir sampling ensures every streaming data point has the same probability, ${mem_{sz}}/{n}$, to be stored in the memory buffer, where $mem_{sz}$ is the size of the buffer and $n$ is the number of data points observed up to now. We summarize the detail in Algorithm~\ref{alg:reservoir} in~\ref{app:algo}. For \modelupdate{}, ER simply trains the model with the incoming and memory mini-batches together using the cross-entropy loss. Despite its simplicity, recent research has shown that ER outperforms many specifically designed CL approaches with and without a memory buffer~\cite{tiny}.

\paragraph{\bf Maximally Interfered Retrieval (MIR)} MIR~\cite{mir} is a recently proposed replay-based method aiming to improve the \retrieval{} strategy. MIR chooses replay samples according to the loss increases given the estimated parameter update based on the incoming mini-batch. Concretely, when receiving a mini-batch $B_n$, MIR performs a virtual parameter update $\theta^v \leftarrow\text{SGD}(B_n ,\theta)$. Then it retrieves the top-k samples from the memory buffer with the criterion $s(x)=l\left(f_{\theta^{v}}(x), y\right) - l\left(f_{\theta}(x), y\right)$, where $x \in M$ and $M$ is the memory buffer. Intuitively, MIR selects memory samples that are maximally interfered (the largest loss increases) by the parameter update with the incoming mini-batch. MIR applies reservoir sampling in \update{} and replays the selected memory samples with new samples in \modelupdate{}. 

\paragraph{\bf Gradient based Sample Selection (GSS)} GSS~\cite{gss} is another replay-based method that focuses on the  \update{} strategy\footnote{\cite{gss} proposes two gradient-based methods, and we select the more efficient one with better performance, dubbed GSS-Greedy}. Specifically, it tries to diversify the gradient directions of the samples in the memory buffer. To this end, GSS maintains a score for each sample in the buffer, and the score is calculated by the maximal cosine similarity in the gradient space between the sample and a random subset from the buffer. When a new sample arrives and the memory buffer is full, a randomly selected subset is used as the candidate set for replacement. The score of a sample in the candidate set is compared to the score of the new sample, and the sample with a lower score is more likely to be stored in the memory buffer. 
Algorithm~\ref{alg:GSS} in \ref{app:algo} shows the main steps of this update method. Same as ER, GSS uses random sampling in \retrieval{}.

\paragraph{\bf Greedy Sampler and Dumb Learner (GDumb)}GDumb~\cite{prabhu2020gdumb} is not specifically designed for CL problems but shows very competitive performance. Specifically, it greedily updates the memory buffer from the data stream with the constraint to keep a balanced class distribution (Algorithm~\ref{alg:gdumb} in~\ref{app:algo}). At inference, it trains a model from scratch using the balanced memory buffer only.

\subsection{Parameter-isolation-based methods}
\paragraph{\bf Continual Neural Dirichlet Process Mixture (CN-DPM)}CN-DPM~\cite{dirichlet} is one of the first dynamic architecture methods that does not require a task-ID. The intuition behind this method is that if we train a new model for a new task and leave the existing models intact, we can retain the knowledge of the past tasks. Specifically, CN-DPM is comprised of a group of experts, where each expert is responsible for a subset of the data and the group is expanded based on the Dirichlet Process Mixture~\cite{ferguson1983bayesian} with Sequential Variational Approximation~\cite{lin2013online}. Each expert consists of a discriminative model (classifier) and a generative model (VAE~\cite{kingma2013auto} is used in this work). The goal of CN-DPM is to model the overall conditional distribution as a mixture of task-wise conditional distributions as the following, where $K$ is the number of experts in the current model: 

\begin{ceqn}
\begin{align}
p(y\mid x)=\sum_{k=1}^{K} \underbrace{p(y \mid x, z=k)}_{\text {discriminative }} \frac{\overbrace{p(x \mid z=k)}^{\text {generative }} {p(z=k)}}{\sum_{k^{\prime}=1}^{K} p\left(x \mid z=k^{\prime}\right) p\left(z=k^{\prime}\right)}
\end{align}
\end{ceqn}

\section{Tricks for Memory-Based Methods in the OCI Setting}
\label{trick}


In class incremental learning, old class samples are generally not available while training on new class samples. Although keeping a portion of old class samples in a memory buffer has been proven effective \cite{tiny, rebuffi2017icarl}, the \textit{class imbalance} is still a serious problem given a limited buffer size.  Moreover, multiple recent works have revealed that class imbalance is one of the most crucial causes of catastrophic forgetting \cite{EEIL, wu2019large, unified}. To alleviate the class imbalance, many simple but effective tricks have been proposed as the building blocks for CL methods by modifying the loss function, post-processing, or using different types of classifiers. 


We will disclose how class imbalance results in recency bias, which has been identified as the main source of catastrophic forgetting in Section~\ref{biased_fc}. Section~\ref{trick_detail} will introduce the compared tricks in detail and explain how they alleviate recency bias.

\subsection{Biased FC layer}
\label{biased_fc}
To better understand how class imbalance affects the learning performance, we define the following notations. The convolutional neural network (CNN) classification model can be split into a feature extractor $\phi(\cdot):\mathbf{x} \mapsto \mathbb{R}^{d}$ where d is the dimension of the feature vector of image $\mathbf{x}$, and a fully-connected (FC) layer with Softmax output. The output of the FC layer is obtained with:

\begin{ceqn}
\begin{align}
logits(\mathbf{x}) = \mathbf{W}^{T}\phi(\mathbf{x})\label{eq: logit}\\
\mathbf{o}(\mathbf{x})=Softmax(logits(\mathbf{x})))
\end{align}
\end{ceqn}
where $\mathbf{W}\in \mathbb{R}^{d\times({\abs{C_{old}} + \abs{C_{new}}}})$, and $C_{old}$ and $C_{new}$ are the sets of old and new classes respectively, with $|\cdot|$ denoting the number of classes in each set. $\mathbf{W_{i}}$ represents the weight vector for class i. For notational brevity, $\mathbf{W}$ also contains the bias terms.

A recent work reveals how does class imbalance result in a biased $\mathbf{W}$~\cite{SS}. We denote $s_i = \mathbf{W_{i}}^{T}\phi(\mathbf{x})$ as the logit of sample $\mathbf{x}$ for class $i$. The gradient of the cross-entropy loss $L_{CE}$ with respect to $\mathbf{W_{i}}$ for the Softmax output is:


\begin{ceqn}
\begin{align}
\frac{\partial \mathcal{L}_{\mathrm{CE}}}{\partial \mathbf{W_{i}}}=\bigg(\frac{e^{s_{i}}}{\sum_{j\in C_{old} + C_{new}} e^{s_{j}}}-\mathbbm{1}_{\{i=y\}}\bigg)\phi(\mathbf{x})
\end{align}
\end{ceqn}
where $y$ is the ground-truth class and  $\mathbbm{1}_{\{i=y\}}$ is the indicator for $i=y$. Since ReLU is often used as the activation function for the embedding networks~\cite{resnet18}, $\phi(\mathbf{x})$ is always positive. Therefore, the gradient is always positive for $i\ne y$. If $i$ belongs to old classes, $i\ne y$ will hold most of the time as the new class samples significantly outnumber the old class samples during training. Thus, the logit for class $i$ will keep being penalized during the gradient descent. As a result, the logits for the old classes are prone to be smaller than those for the new classes, and the model is consequently biased towards new classes. 

We will perform quantitative error analysis in Section~\ref{error_ana} and empirically show that the logits for old classes are significantly smaller than the logits for new classes. 

Other than the bias related to the Softmax classifier mentioned above, another problem induced by class imbalance is under-representation of the minority~\cite{dong2018imbalanced, yin2018feature}, where minority classes do not show a discernible pattern in the latent feature space~\cite{prs}. The under-representation introduces additional difficulty for other classifiers apart from the Softmax classifier, such as nearest-class-mean~\cite{ncm} and cosine-similarity-based classifier~\cite{gidaris2018dynamic}.

\subsection{Compared tricks}
\label{trick_detail}
To mitigate the strong bias towards new classes due to class imbalance, multiple methods have been proposed lately~\cite{lee2019overcoming, jung2018less, wa, il2m, douillard2020podnet}. For example, LUCIR~\cite{unified} proposes three tricks: cosine normalization to balance class magnitudes, a margin ranking loss for inter-class separation, and a less-forgetting constraint to preserve the orientation of features. BiC~\cite{wu2019large} trains an additional linear layer to remove bias with a separate validation set. We compare seven simple (effortless integration without additional resources) but effective (decent improvement) tricks in this work.

\paragraph{\bf Labels Trick (LB)}~\cite{LB} proposes to consider only the outputs for the classes in the current mini-batch when calculating the cross-entropy loss, in contrast to the common practice of considering outputs for all the classes. To achieve this, the outputs that do not correspond to the classes of the current mini-batch are masked out when calculating the loss.


Although the author did not demonstrate the motivation of this trick, we can easily find the rationale based on the analysis in Section~\ref{biased_fc}. Masking out all the outputs that don't match the classes in the current mini-batch is equivalent to changing the loss function to:

\begin{ceqn}
\begin{align}
\mathcal{L}_{\mathrm{CE}}(x_i, y_i) = -\log \left(\frac{e^{s_{y_{i}}}}{\sum_{j \in {C}_{cur}} e^{s_{j}}}\right)
\end{align}
\end{ceqn}
where $C_{cur}$ denotes the classes in the current mini-batch. We can see that $\frac{\partial \mathcal{L}_{\mathrm{CE}}}{\partial s_{j}}=0$ for $j \notin C_{cur}$, and therefore training with the current mini-batch will not overly penalize the logits for classes that are not in the mini-batch.

\paragraph{\bf Knowledge Distillation with Classification Loss (KDC)}KD~\cite{distill} is an effective way for knowledge transfer between networks. Multiple recent works have proposed different ways to combine the KD loss with the classification loss~\cite{rebuffi2017icarl, EEIL, javed2018revisiting}. In this part, we compare the methods from~\cite{wu2019large}. Specifically, the loss function is given as:
\begin{ceqn}
\begin{align}
\mathcal{L}(\mathbf{x}, y)=\lambda \mathcal{L}_{C E}(\mathbf{x}, y)+ (1-\lambda)\mathcal{L}_{K D}(\mathbf{x})
\end{align}
\end{ceqn}
where $\lambda$ is set to $\frac{\abs{C_{new}}}{\abs{C_{old}}+\abs{C_{new}}}$. Note that $(\mathbf{x}, y)$ is from both new class data from the current task and old class data from the memory buffer. 

As shown in Table~\ref{table:trick_cifar}, however, this method does not perform well in our experiment setting, especially with a large memory buffer. We identify $\lambda$ as the key issue. We find that the accuracy for new class samples becomes almost zero around the end of training because $\lambda$ is very small. In other words, $\mathcal{L}_{K D}$ dominates the loss, and the model cannot learn any new knowledge. Hence, we suggest setting $\lambda$ to $\sqrt{\frac{\abs{C_{new}}}{\abs{C_{old}}+\abs{C_{new}}}}$. We denote the trick with this modification as KDC*. 

\paragraph{\bf Multiple Iterations (MI)}
Most of the previous works only perform a single gradient update on the incoming mini-batch in the online setup. \cite{mir} suggests performing multiple gradient updates to maximally utilize the current mini-batch. Particularly for replay methods, additional updates with different replay samples can improve performance. We run 5 iterations per incoming mini-batch and retrieve different memory samples for each iteration in this work.

\paragraph{\bf Nearest Class Mean (NCM) Classifier}To tackle the biased FC layer, one can replace the FC layer and Softmax classifier with another type of classifier.  Nearest Class Mean classifier (NCM)~\cite{ncm} is a popular option in CL~\cite{rebuffi2017icarl, yu2020semantic}. To make prediction for a sample $\mathbf{x}$, NCM computes a prototype vector for each class and assigns the class label with the most similar prototype:
\begin{ceqn}
\begin{align}
\mu_{y}=\frac{1}{\left|M_{y}\right|} \sum_{\mathbf{x}_m \in M_{y}} \phi(\mathbf{x}_m)\label{eq:prototye}\\
y^{*}=\underset{y=1, \ldots, t}{\operatorname{argmin}}\left\|\phi(\mathbf{x})-\mu_{y}\right\|
\end{align}
\end{ceqn}
In the class incremental setting, the true prototype vector for each class cannot be computed due to the unavailability of the training data for previous tasks. Instead, the prototype vectors can be approximated using the data in the memory buffer. In Eq.~\eqref{eq:prototye}, $M_{y}$ denotes the memory samples of class $y$.

\paragraph{\bf Separated Softmax (SS)}
Since training the whole FC layer with one Softmax output layer results in bias as explained in Section~\ref{biased_fc}, SS~\cite{SS} employs two Softmax output layers: one for new classes and another one for old classes. The loss function can be calculated as below:
\begin{ceqn}
\begin{align}
\begin{aligned}
& \mathcal{L}(\mathbf{x}_{i}, y_{i}) \\
=&-\log \left(\frac{e^{s_{y_{i}}}}{\sum_{j \in {C}_{old}} e^{s_{j}}}\right) \cdot \mathbbm{1}\left\{y_{i} \in {C}_{old}\right\}\\
& -\log \left(\frac{e^{s_{y_{i}}}}{\sum_{j \in {C}_{new}} e^{s_{j}}}\right) \cdot \mathbbm{1}\left\{y_{i} \in {C}_{new}\right\}
\end{aligned}
\end{align}
\end{ceqn}
Depending on whether $y_i \in C_{new}$ or $C_{old}$, the corresponding Softmax is used to compute the cross-entropy loss. We can find that $\frac{\partial \mathcal{L}}{\partial s_{j}}=0$ for $j \in C_{old}$ and $y_i \in C_{new}$. Thus, training with new class samples will not overly penalize the logits for the old classes.

\paragraph{\bf Review Trick (RV)}
To alleviate the class imbalance, \cite{EEIL} proposes an additional fine-tuning step with a small learning rate, which uses a balanced subset from the memory buffer and the training set of the current task. A temporary distillation loss for new classes is applied to avoid forgetting the new classes during the fine-tuning phase mentioned above. A similar yet simplified version, dubbed Review Trick, is applied in the winning solution in the continual learning challenge at CVPR2020~\cite{zheda}. At the end of learning the current task, the review trick fine-tunes the model with all the samples in the memory buffer using only the cross-entropy loss. In this work, we compare the review trick from~\cite{zheda} with a learning rate 10 times smaller than the training learning rate. 

\begin{table*}[t!]
\centering
\scriptsize
\begin{tabular}{l | c c c c c c}
    
    \toprule
    Dataset &\#Task&\#Train/task&\#Test/task&\#Class&Image Size&Setting\Bstrut \\ 
    \hline\hline
    \Tstrut
    Split MiniImageNet& 20&2500&500&100&3x84x84&OCI\\
    Split CIFAR-100&20&2500&500&100&3x32x32&OCI\\
    CORe50-NC&9&12000$\sim$24000&4500$\sim$9000&50&3x128x128&OCI\\
    NS-MiniImageNet&10&5000&1000&100&3x84x84&ODI\\
    CORe50-NI\footnotemark&8&15000&44972&50&3x128x128&ODI\\
    \bottomrule
\end{tabular}

\caption{Summary of dataset statistics}
\label{tab:dataset}
\end{table*}

\section{Online Continual Hyperparameter Tuning}
\label{hyperparameter}
\begin{algorithm*}
\footnotesize
\caption{Hyperparameter Tuning Protocol}
\label{alg:tune}
\SetCustomAlgoRuledWidth{0.49\textwidth}
\SetAlgorithmName{Algorithm}{}{}
    \SetKwInOut{Input}{Input~}
    \SetKwInput{Require}{Require}
\Input{Hyperparameter set $\mathcal{P}$}
\Require{$D^{CV}$ data stream for tuning, $D^{EV}$ data stream for learning \& testing}
\Require{$f$ classifier, $A^{CL}$ CL algorithm}

\For(\Comment{Multiple passes over $D^{CV}$ for tuning}){$p \in \mathcal{P}$}{
    \For(\Comment{Single pass over $D^{CV}$ with $p$}){$i\in\{1,\dots,T^{CV}\}$}{ \
        \For{$B_n\sim D^{CV}_i$}{
        $A^{CL}(f, B_n, p)$
        }
        $\text{Evaluate}(f, D^{CV}_{test})$        \Comment{Store performance on test set of $D^{CV}$}
    }
}
$\text{Best hyperparameters, } \pmb{p^*} \leftarrow$ based on Average Accuracy of $D^{CV}_{test}$, see Eq.\eqref{eq: acc}

 \For(\Comment{Learning over $D^{EV}$}){$i\in\{1,\dots,T^{EV}\}$}{
        \For(\Comment{Single pass over $D^{EV}$ with $p^*$}){$B_n\sim D^{EV}_i$}{
        $A^{CL}(f, B_n, \pmb{p^*})$
        }
        $\text{Evaluate}(f, D^{EV}_{test})$
}
Report performance on $D^{EV}_{test}$
\end{algorithm*}
In practice, most CL methods have to rely on well-selected hyperparameters to effectively balance the trade-off between stability and plasticity. Hyperparameter tuning, however, is already a challenging task in learning conventional deep neural network models, which becomes even more complicated in the online CL setting. Meanwhile, a large volume of CL works still tune hyperparameters in an offline manner by sweeping over the whole data sequence and selecting the best hyperparameter set with grid-search on a validation set.
After that, metrics are reported on the test set with the selected set of hyperparameters. This tuning protocol violates the online CL setting where a classifier can only make a single pass over the data, which implies that the reported results in the CL literature may be too ideal and cannot be reproduced in real online CL applications. 

Recently, several hyperparameter tuning protocols that are useful for CL settings have been proposed. \cite{comprehensive} introduces a tuning protocol for two tasks ($D_1$ and $D_2$). Firstly, they determine the best combination of model hyperparameters using $D_1$. Then, they tune the learning rate to be used for learning $D_2$ such that the test accuracy on $D_2$ is maximized. \cite{survey} proposes another protocol that dynamically determines the stability-plasticity trade-off. When the model receives a new task, the hyperparameters are set to ensure minimal forgetting of previous tasks. If a predefined threshold for the current task's performance is not met, the hyperparameters are adjusted until achieving the threshold. While the aforementioned protocols assume the data of a new task is available all at once, \cite{agem} presents a protocol targeting the online CL. Specifically, a data stream is divided into two sub-streams --- $D^{CV}$, the stream for cross-validation and $D^{EV}$, the stream for final training and evaluation. Multiple passes over $D^{CV}$ are allowed for tuning, but a CL algorithm can only perform a single pass over $D^{EV}$ for training. The metrics are reported on the test sets of $D^{EV}$. Since the setting of our interest in this work is the online CL, we adopt the protocol from~\cite{agem} for our experiments. We summarize the protocol in Algorithm~\ref{alg:tune}. 

\section{Experiments}
\label{experiments}
Section~\ref{exp_setup} explains the general setting for all experiments. Then, we focus on the OCI setting in Section~\ref{perf_class_incremental} and \ref{performance_tricks}: we evaluate all the compared methods and baselines in Section~\ref{perf_class_incremental} and investigate the effectiveness of the seven tricks in Section~\ref{performance_tricks}. In Section~\ref{perf_domain_incremental}, we assess the compared methods in the ODI setting to investigate their abilities to generalize, and Section~\ref{comment} provides general comments on the surveyed methods and tricks. 


\footnotetext{CORe50-NI uses one test set(44972 images) for all tasks}
\subsection{Experimental setup}
\label{exp_setup}
\paragraph{Datasets}
We evaluate the nine methods summarized in Section~\ref{methods} and additional two baselines on three class incremental datasets.  We also propose a new domain incremental dataset based on Mini-ImageNet and examine the compared methods in the ODI setting to see how well the methods can generalize to this setting. The summary of dataset statistics is provided in Table~\ref{tab:dataset}.

\paragraph{Class Incremental Datasets}
\begin{itemize}
  \item \textbf{Split CIFAR-100} is constructed by splitting the CIFAR-100 dataset~\cite{cifar} into 20 tasks with disjoint classes, and each task has 5 classes. There are 2,500 3\texttimes 32\texttimes 32 images for training and 500 images for testing in each task.
  \item \textbf{Split MiniImageNet} splits MiniImageNet dataset \cite{miniimage}, a subset of  ImageNet~\cite{deng2009imagenet} with 100 classes, into 20 disjoint tasks as in~\cite{tiny}. Each task contains 5 classes, and every class consists of 500 3\texttimes 84\texttimes 84 images for training and 100 images for testing. 
  \item \textbf{CORe50-NC}~\cite{lomonaco2017core50} is a benchmark designed for class incremental learning with 9 tasks and 50 classes: 10 classes in the first task and 5 classes in the subsequent 8 tasks. Each class has around 2,398 3\texttimes 128\texttimes 128 training images and 900 testing images. 
\end{itemize}

\paragraph{Domain Incremental Datasets}
\begin{itemize}
  \item \textbf{NonStationary-MiniImageNet} (NS-MiniImageNet) The most popular domain incremental datasets are still based on MNIST~\cite{mnist}, such as Rotation MNIST~\cite{gem} and Permutation MNIST~\cite{Kirkpatrick2017}. To evaluate the domain incremental setting in a more practical scenario, we propose NS-MiniImageNet with three nonstationary types: noise, blur and occlusion. The number of tasks and the strength of each nonstationary type are adjustable in this dataset. In this survey, we use 10 tasks for each type, and each task comprises 5,000 3\texttimes 84\texttimes 84 training images and 1,000 testing images. As shown in Table~\ref{tab: ns_mini}, the nonstationary strength increases over time, and to ensure a smooth distribution shift, the strength always increases by the same constant. More details about the nonstationary strengths used in the experiments can be found in \ref{app:exp_detail}. 
  

  \item \textbf{CORe50-NI}~\cite{lomonaco2017core50} is a practical benchmark designed for assessing the domain incremental learning with 8 tasks, where each task has around 15,000 training images of 50 classes with different types of nonstationarity including illumination, background, occlusion, pose and scale. There is one single test set for all tasks, which contains 44,972 images. 
\end{itemize}

\begin{table}[h!]
\scriptsize
\begin{center}
\begin{tabular}{|c|c|c|c|c|c|}
\hline
Type&Task 1&...&Task 5&...&Task 10\\ \hline
Noise
&
\raisebox{-.4\height}{\includegraphics[width=0.08\textwidth ]{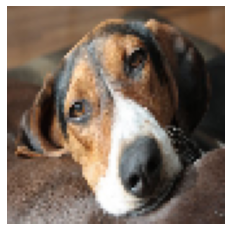}}&
...&
\raisebox{-.4\height}{\includegraphics[width=0.08\textwidth ]{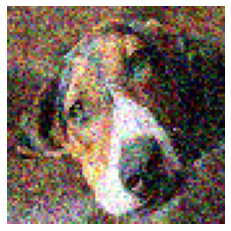}}&
...&
\raisebox{-.4\height}{\includegraphics[width=0.08\textwidth ]{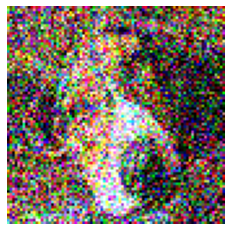}}
\\

\hline
Blur
&

\raisebox{-.4\height}{\includegraphics[width=0.08\textwidth]{pics/dog_ns.png}}&
...&
\raisebox{-.4\height}{\includegraphics[width=0.08\textwidth]{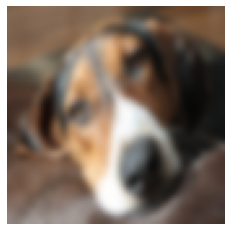}}&
...&
\raisebox{-.4\height}{\includegraphics[width=0.08\textwidth]{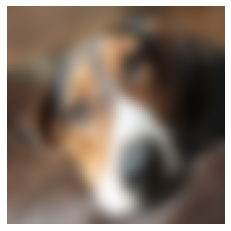}}

\\\hline

Occlusion
&

\raisebox{-.4\height}{\includegraphics[width=0.08\textwidth]{pics/dog_ns.png}}&
...&
\raisebox{-.4\height}{\includegraphics[width=0.08\textwidth]{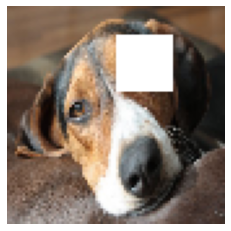}}&
...&
\raisebox{-.4\height}{\includegraphics[width=0.08\textwidth]{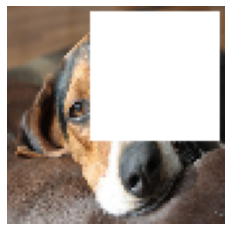}}\\
\hline
\end{tabular}
\caption{Example images of different nonstationary types in NS-MiniImageNet. The nonstationary strength increases over time, and to ensure a smooth distribution shift, the strength always increases by the same constant.}
\label{tab: ns_mini}
\end{center}
\end{table}
\paragraph{Task Order and Task Composition}
Since the task order and task composition may impact the performance~\cite{lomonaco2017core50}, we take the average over multiple runs for each experiment with different task orders and composition to reliably assess the robustness of the methods. 
For CORe50-NC and CORe50-NI, we follow the number of runs (i.e., 10), task order and composition provided by the authors. For Split CIFAR-100 and Split MiniImagenet, we average over 15 runs, and the class composition in each task is randomly selected for each run.

\begin{table*}[t!]
\tiny
\resizebox{\textwidth}{!}{
\begin{tabular}{l|ccc|ccc|ccc}
\toprule
Method& \multicolumn{3}{c|}{Split CIFAR-100}&\multicolumn{3}{c|}{Split Mini-ImageNet}&\multicolumn{3}{c}{CORe50-NC}\\
\midrule
Finetune & \multicolumn{3}{c|}{$3.7\pm0.3$}&\multicolumn{3}{c|}{$3.4\pm0.2$}&\multicolumn{3}{c}{$7.7\pm1.0$}\\
Offline & \multicolumn{3}{c|}{$49.7\pm2.6$}&\multicolumn{3}{c|}{$51.9\pm0.5$}&\multicolumn{3}{c}{$51.7\pm1.8$}\\
EWC++ & \multicolumn{3}{c|}{$3.7\pm0.4$}&\multicolumn{3}{c|}{$3.5\pm0.4$}&\multicolumn{3}{c}{$8.3\pm0.3$}\\
LwF & \multicolumn{3}{c|}{$7.2\pm0.4$}&\multicolumn{3}{c|}{$7.6\pm0.7$}&\multicolumn{3}{c}{$7.1\pm1.9$}\\
\midrule
Buffer Size &M=1k& M=5k &  M=10k &M=1k& M=5k &  M=10k&M=1k& M=5k &  M=10k\\
\midrule
ER &$7.6\pm0.5$&$17.0\pm1.9$&$18.4\pm1.4$&$6.4\pm0.9$&$14.5\pm2.1$&$15.9\pm2.0$&$23.5\pm2.4$&$27.5\pm3.5$&$28.2\pm3.3$\\
MIR &$7.6\pm0.5$&$18.2\pm0.8$&$19.3\pm0.7$&$6.4\pm0.9$&$16.5\pm2.1$&$21.0\pm1.1$&$\mathbf{27.0\pm1.6}$&$\mathbf{32.9\pm1.7}$&$\mathbf{34.5\pm1.5}$\\
GSS &$7.7\pm0.5$&$11.3\pm0.9$&$13.4\pm0.6$&$5.9\pm0.7$&$11.2\pm0.9$&$13.5\pm0.8$&$19.6\pm3.0$&$22.2\pm4.4$&$21.1\pm3.5$\\
iCaRL &$\mathbf{16.7\pm0.8}$&$19.2\pm1.1$&$18.8\pm0.9$&$\mathbf{14.7\pm0.4}$&$17.5\pm0.6$&$17.4\pm1.5$&$22.1\pm1.4$&$25.1\pm1.6$&$22.9\pm3.1$\\
A-GEM &$3.7\pm0.4$&$3.6\pm0.2$&$3.8\pm0.2$&$3.4\pm0.2$&$3.7\pm0.3$&$3.3\pm0.3$&$8.7\pm0.6$&$9.0\pm0.5$&$8.9\pm0.6$\\
CN-DPM &$14.0\pm1.7$&-&-&$9.4\pm1.2$&-&-&$7.6\pm0.4$&-&-\\
GDumb &$10.4\pm1.1$&$\mathbf{22.1\pm0.9}$&$\mathbf{28.8\pm0.9}$&$8.8\pm0.4$&$\mathbf{21.1\pm1.7}$&$\mathbf{31.0\pm1.4}$&$15.1\pm1.2$&$28.1\pm1.4$&$32.6\pm1.7$\\
\bottomrule
\end{tabular}}
\caption{Average accuracy (end of training) for the OCI setting of Split CIFAR-100, Split Mini-ImageNet and CORe50-NC. Replay-based methods and a strong baseline GDumb show competitive performance across three datasets.}
\label{table:nc}
\end{table*}

\paragraph{Models}
Similar to \cite{tiny, gem}, we use the reduced ResNet18~\cite{he2016deep} as the base model for all datasets and methods. The network is trained via the cross-entropy loss with a stochastic gradient descent optimizer and a mini-batch size of 10. The size of the mini-batch retrieved from the memory buffer is also set to 10, irrespective of the size of the memory buffer as in~\cite{tiny}. Note that with techniques such as transfer learning (e.g., using a pre-trained model from ImageNet), data augmentation and deeper network architectures, it is possible to achieve much higher performance in this setting~\cite{clvision}. However, since those techniques are orthogonal to our investigation and deviate from the simpler experimental settings of other papers we cite and compare, we do not use them in our experiments.
\paragraph{Baselines}We compare the methods we discussed in Section~\ref{methods} with two baselines:
\begin{itemize}
\item \textbf{Finetune} greedily updates the model with the incoming mini-batch without considering the previous task performance. The model suffers from catastrophic forgetting and is regarded as the lower-bound. 
\item \textbf{Offline} trains a model using all the samples in a dataset in an offline manner. The baseline is trained for multiple epochs within each of which mini-batches are sampled i.i.d from differently shuffled dataset. We train the model for 70 epochs with the mini-batch size of 128. 
\end{itemize}

\paragraph{Other Details} We evaluate the performance with 5 metrics we described in Section~\ref{metrics}: Average Accuracy, Average Forgetting, Forward Transfer, Backward Transfer and Run Time. We use the hyperparameter tuning protocol described in Section~\ref{hyperparameter} and give each method similar tuning budget. The details of the implementation including hyperparameter selection can be found in~\ref{app:impl_detail}.

\subsection{Performance comparison in Online Class Incremental (OCI) setting}
\label{perf_class_incremental}

\begin{figure*}
    \begin{subfigure}{0.32\textwidth}
        \centering
        \includegraphics[
        height=4cm
        ]{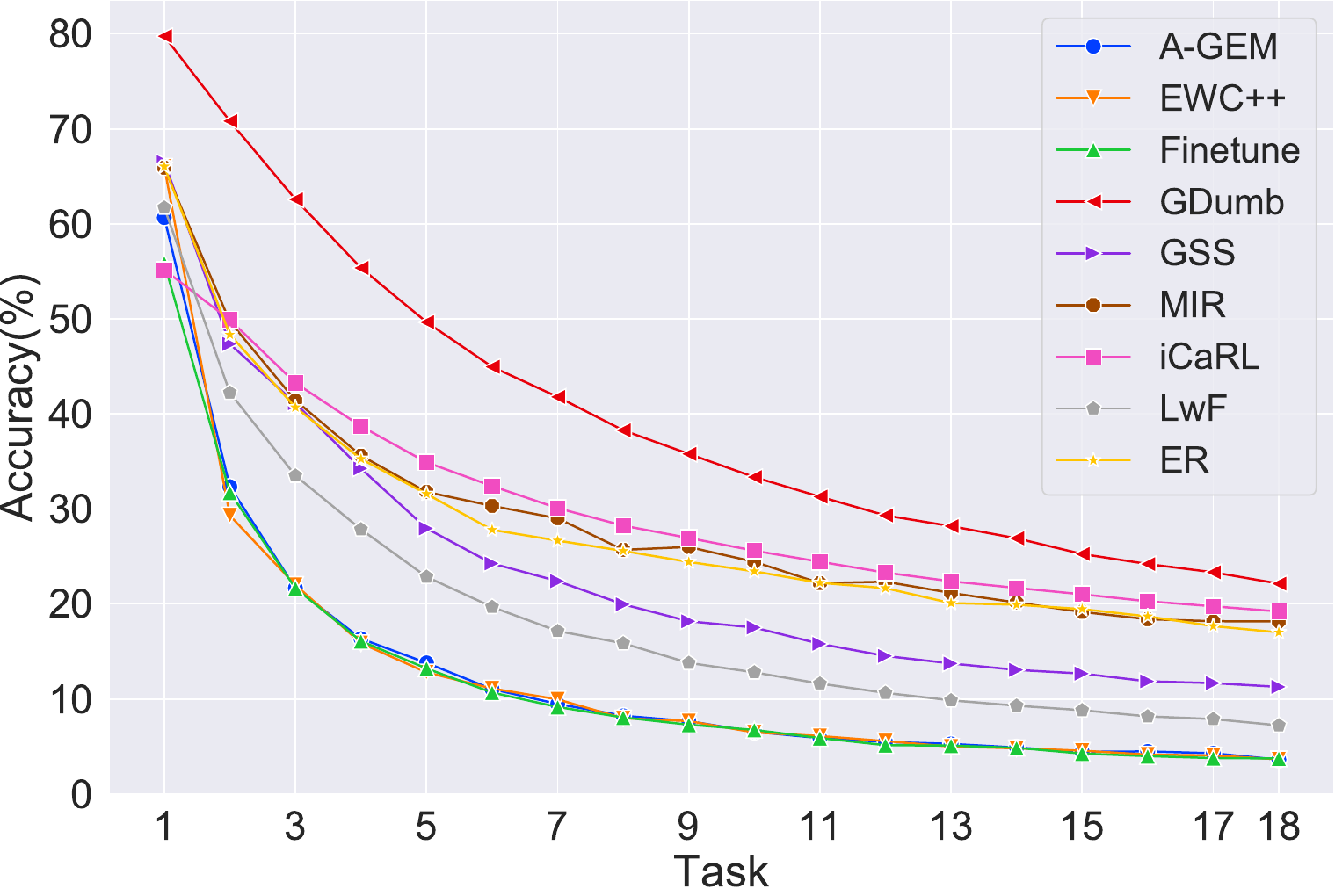}
        \caption{CIFAR-100}
        \label{fig:cifar100_nc}
    \end{subfigure}
    \hfill
    \begin{subfigure}{0.32\textwidth}
    \centering
    \includegraphics[
    height=4cm
    ]{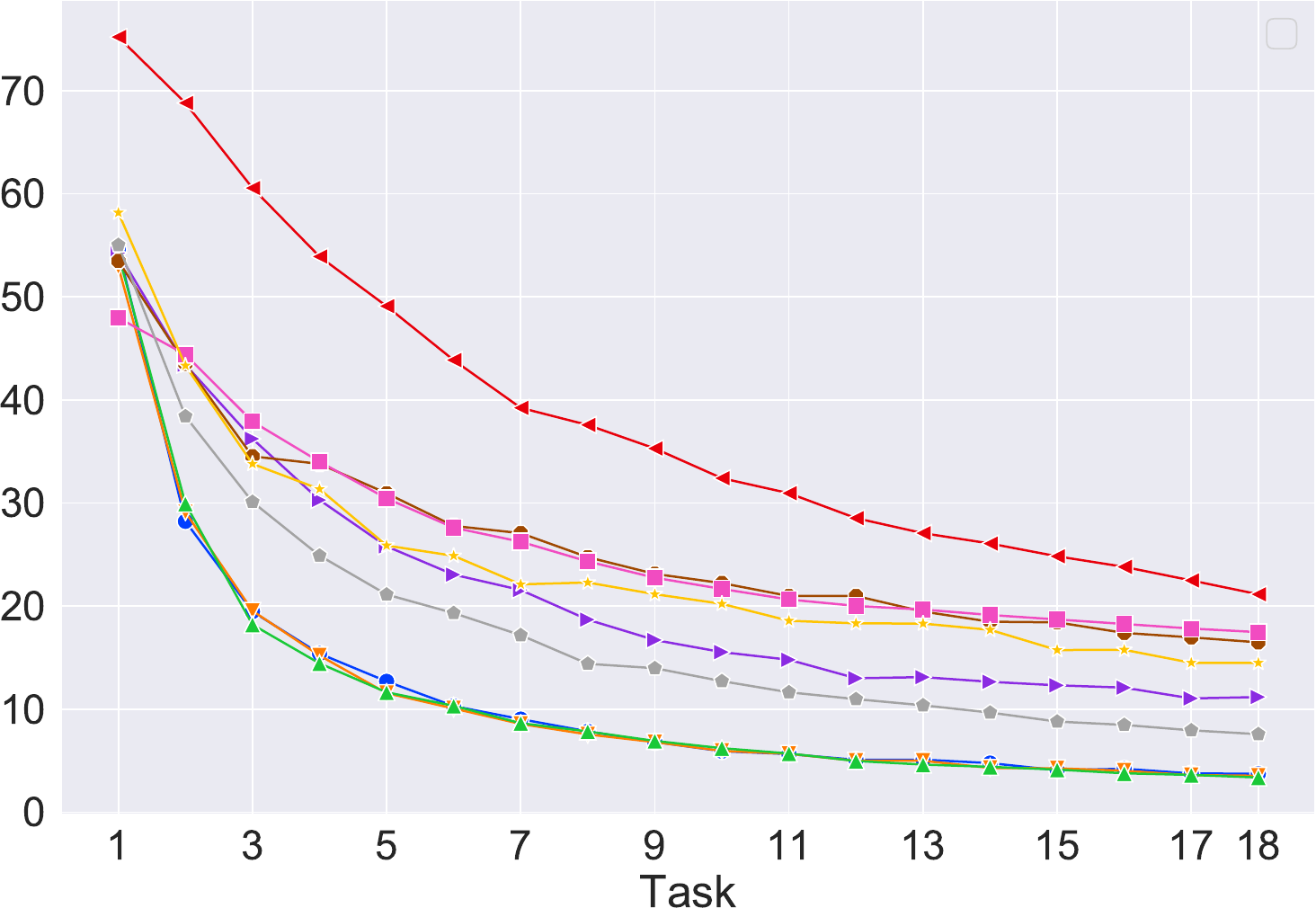}
    \caption{Mini-ImageNet}
    \label{fig:mini_nc}
    \end{subfigure}
    \hfill
    \begin{subfigure}{0.32\textwidth}
    \centering
    \includegraphics[height=4cm
    ]{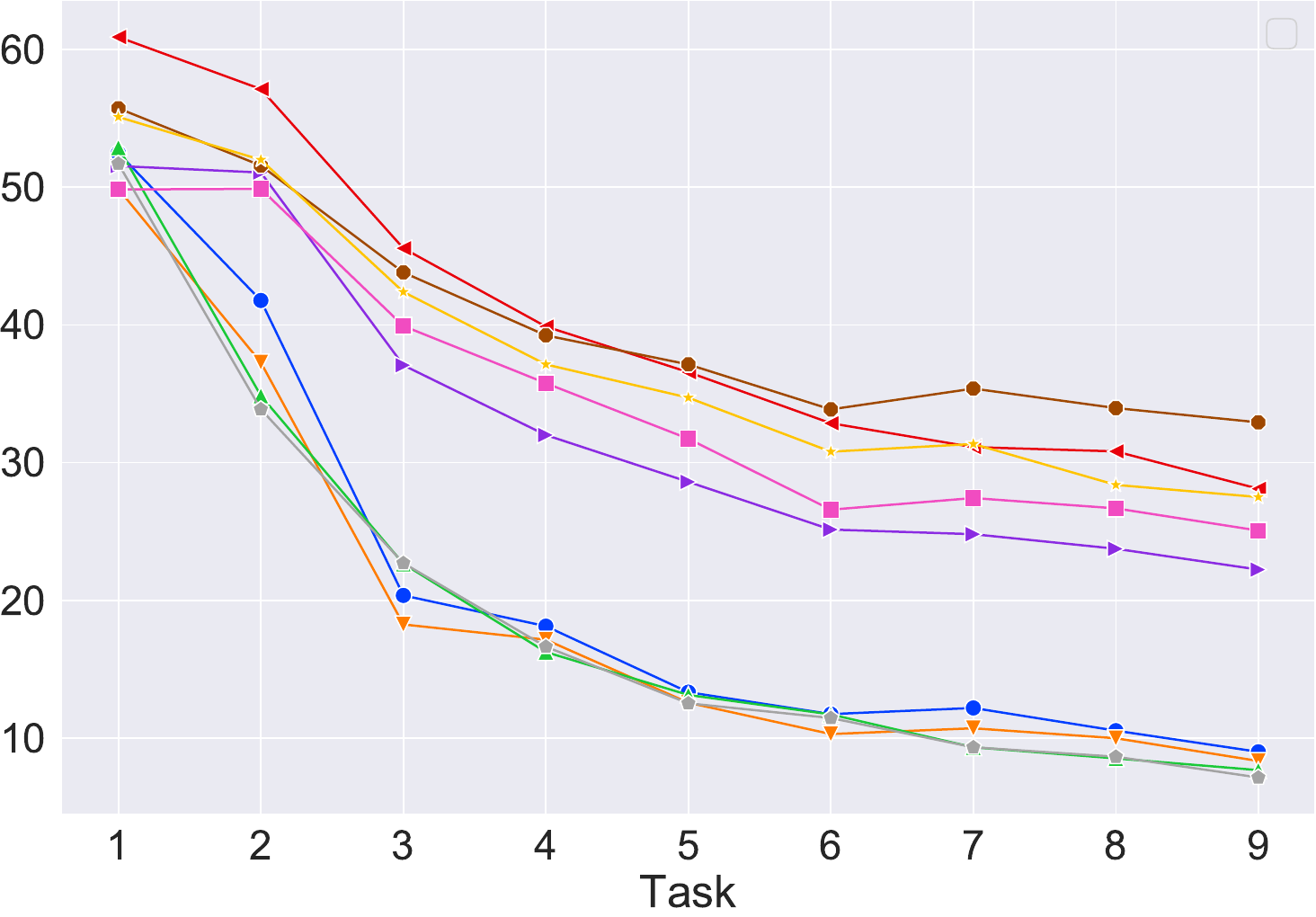}
    \caption{CORe50-NC}
    \label{fig:core_nc}
    \end{subfigure}
    \caption{The average accuracy measured by the end of each task for the OCI setting with a 5k memory buffer. More detailed results for different memory buffer sizes are shown in~\ref{additional_oci}.}\label{fig:nc_5k}
\end{figure*}

\paragraph{Regularization-based Methods}As shown in Table~\ref{table:nc},  EWC++ has almost the same performance as Finetune in the OCI setting. We find that the gradient explosion of the regularization terms is the root cause. Specifically, $\lambda$ in EWC++ controls the regularization strength, and we need a larger $\lambda$ to avoid forgetting. However, when $\lambda$ increases to a certain value, the gradient explosion occurs. If we take $\theta_j$ in Eq.~\eqref{eq:ewc} as an example, the regularization term for $\theta_j$ has the gradient $ {\lambda}F_{j}(\theta_{j}-\theta^{*})$. Some model weights change significantly when it receives data with new classes, and therefore the gradients for those weights are prone to explode with a large $\lambda$. The Huber regularization proposed lately could be a possible remedy~\cite{Huber}. Surprisingly, we also observe that LwF, a method relying on KD, has similar performance as replay-based methods with a small memory buffer(1k) such as ER, MIR and GSS in Split CIFAR-100 and even outperforms them in Split Mini-ImageNet. In the larger and more realistic CORe50-NC, however, both EWC++ and LwF fail. This also confirms the results of three recent studies, where \cite{lesort2019regularization} shows the shortcomings of regularization-based approaches in the class incremental setting, \cite{knoblauch2020optimal} theoretically explains why regularization-based methods underperform memory-based methods and \cite{belouadah2020scail} empirically demonstrates that KD is more useful in small-scale datasets.

\paragraph{Memory-based Methods}
Firstly, A-GEM does not work in this setting as it has almost the same performance as Finetune, implying that the indirect use of the memory samples is less efficient than the direct relay in the OCI setting. Secondly, given a small memory buffer in Split CIFAR-100 and Mini-ImageNet, iCaRL---proposed in 2017---shows the best performance. On the other hand, other replay-based methods such as ER, MIR and GSS do not work well because simply replaying with a small memory buffer yields severe class imbalance. When equipped with a larger memory buffer (5k and 10k), GDumb---a simple baseline that trains with the memory buffer only---outperforms other methods by a large margin. Additionally, as shown in Fig.~\ref{fig:cifar100_nc} and Fig.~\ref{fig:mini_nc}, GDumb dominates the average accuracy not only at the end of training but also at any other evaluation points along the data stream. This raises concerns about the progress in the OCI setting in the literature. Next, in the larger CORe50-NC dataset, GDumb is less effective since it only relies on the memory and the memory is smaller in a larger dataset in proportion. MIR is a robust and strong method as it exhibits remarkable performance across different memory sizes. Also, even though GSS is claimed to be an enhanced version of ER, ER consistently surpasses GSS across different memory sizes and datasets, which is also confirmed by other studies~\cite{dirichlet, buzzega2020dark}.

\paragraph{Parameter-isolation-based methods}
As one of the first dynamic-architecture methods without using a task-ID, CN-DPM shows competitive results in Split CIFAR-100 and Mini-ImageNet but fails in CORe5-NC. The key reason is that CN-DPM is very sensitive to hyperparameters, and when applying CN-DPM in a new dataset, a good performance cannot be guaranteed given a limited tuning budget. 
\begin{figure*}
    \centering
    \includegraphics[
    height=13cm]{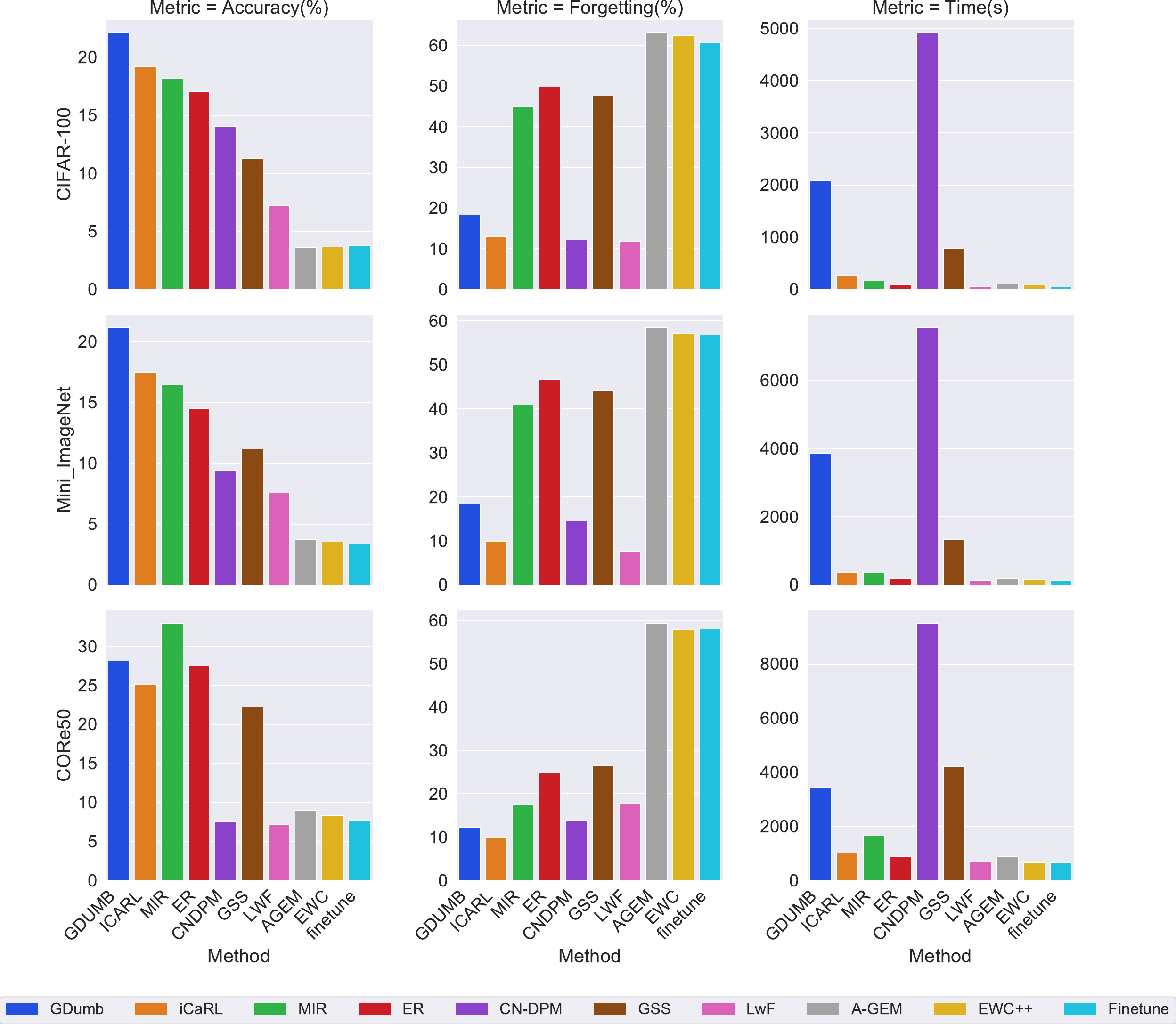}
    \caption{Average accuracy, forgetting and running time for the OCI setting with a 5k memory buffer. Each column represents a metric and each row represents a dataset. Note that in the OCI setting, none of the methods show any forward and backward transfer and hence they are not shown in this figure. }\label{fig:nc_bar}
\end{figure*}

\paragraph{Other Metrics}
We show the performance of all five metrics in Fig.~\ref{fig:nc_bar}. Generally speaking, we find that a high average accuracy comes with low forgetting, but methods using KD such as iCaRL and LwF, and dynamic-architecture methods such as CN-DPM have lower forgetting. The reason for the low forgetting of these methods is intransigence, the model's inability to learn new knowledge~\cite{chaudhry2018riemannian}. For iCaRL and LwF, KD imposes a strong regularization, which may lead to a lower accuracy on new tasks. For CN-DPM, the inaccurate expert selector is the cause of the intransigence~\cite{dirichlet}.  Furthermore, most methods have similar running time except for CN-DPM, GDumb and GSS. CN-DPM requires a significantly longer training time as it needs to train multiple experts, and each expert contains a generative model (VAE~\cite{kingma2013auto}) and a classifier(10-layer ResNet~\cite{he2016deep}). GDumb has the second longest running time as it requires training the model from scratch with the memory at every evaluation point. Lastly, we notice none of the methods show any forward and backward transfer, which is expected since a model tends to classify all the test samples as the current task labels due to the strong bias in the last FC layer.

\subsection{Error analysis of memory-based methods in the OCI setting}
\label{error_ana}

We perform quantitative error analysis for three memory-based methods with 5k memory buffer (A-GEM, ER and MIR) on Split CIFAR-100. We define $e(n, o)$ and $e(n, n)$ as the number of test samples from new classes that are misclassified as old classes and new classes, respectively. The same notation rule is applied to $e(o, o)$ and $e(o, n)$. Also, $er(n, o)$ denotes the ratio of new class test samples misclassified as old classes to the total number of new class test samples. $er(o, n)$ is similarly defined.

\begin{table}[t!]
\scriptsize
\begin{center}
\begin{tabular}{c|cccc|cc}
\toprule
Method & e(n, o) & e(n, n) & e(o, o) & e(o, n)&er(n, o)&er(o, n)\\
\midrule
\midrule
A-GEM & 0 & 177 & 0 & 9500&0\%&100\% \\
ER & 37 &148 & 2269 & 5852&20\%&72\% \\
MIR & 54 & 113 & 2770 & 5330&32\%&66\%\\
\bottomrule
\end{tabular}
\end{center}
\caption{Error analysis of CIFAR-100 by the end of training with M=5k. e(n, o) \& e(n, n) represent the number of test samples from new classes that are misclassified as old classes and new classes, respectively. Same notation rule is applied to e(o, o) \& e(o, n). 
}
\label{table:trick_motivation}
\end{table}

\begin{figure}[t]
    \begin{subfigure}[t]{0.235\textwidth}
        \centering
        \includegraphics[
        height=3.1cm
        ]{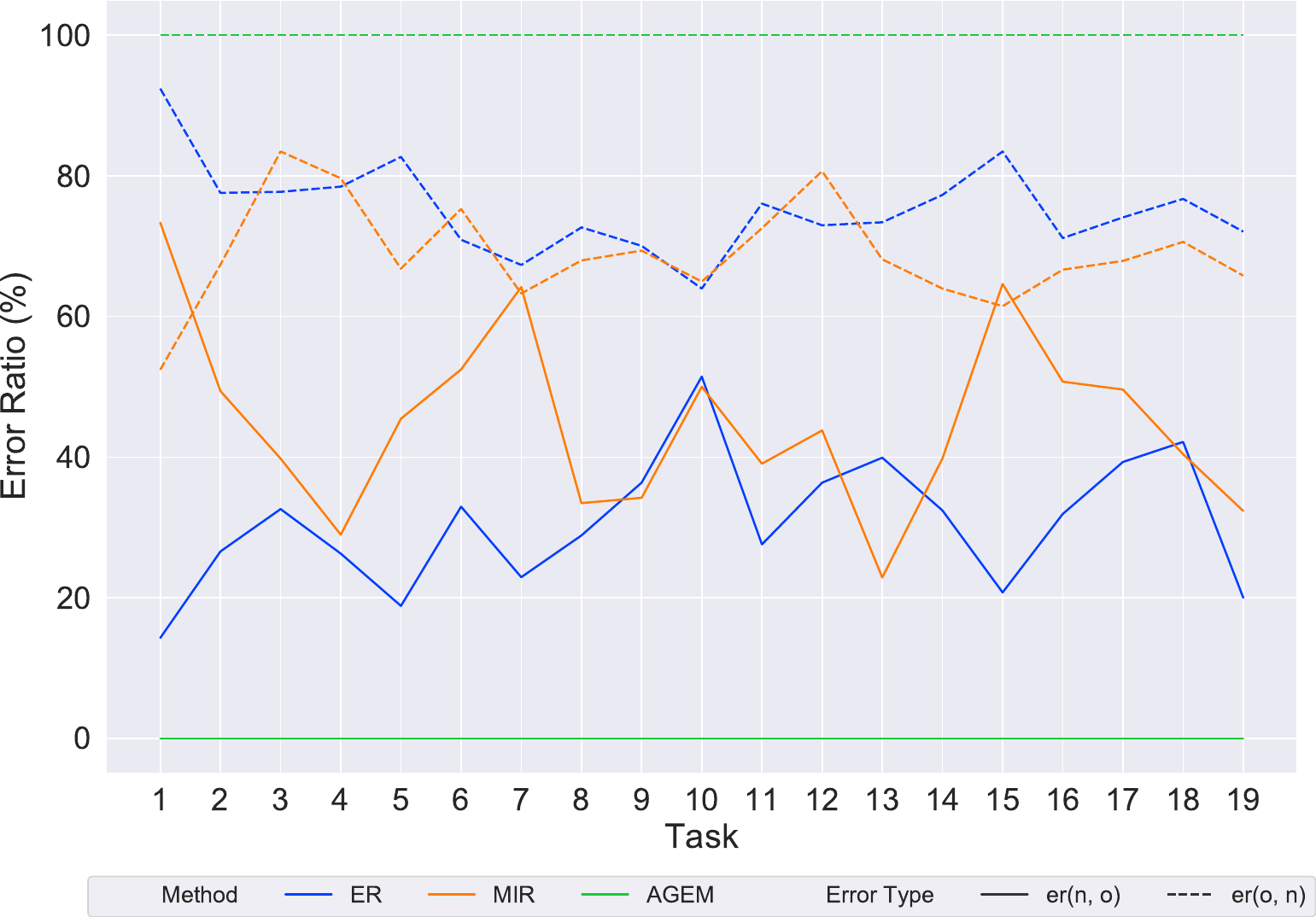}
        \caption{The Error Ratio er(n, o) and er(o, n)}
        \label{fig:trick_motivation_1}
    \end{subfigure}
    \hfill
    \begin{subfigure}[t]{0.235\textwidth}
    \centering
    \includegraphics[
    height=3.1cm
    ]{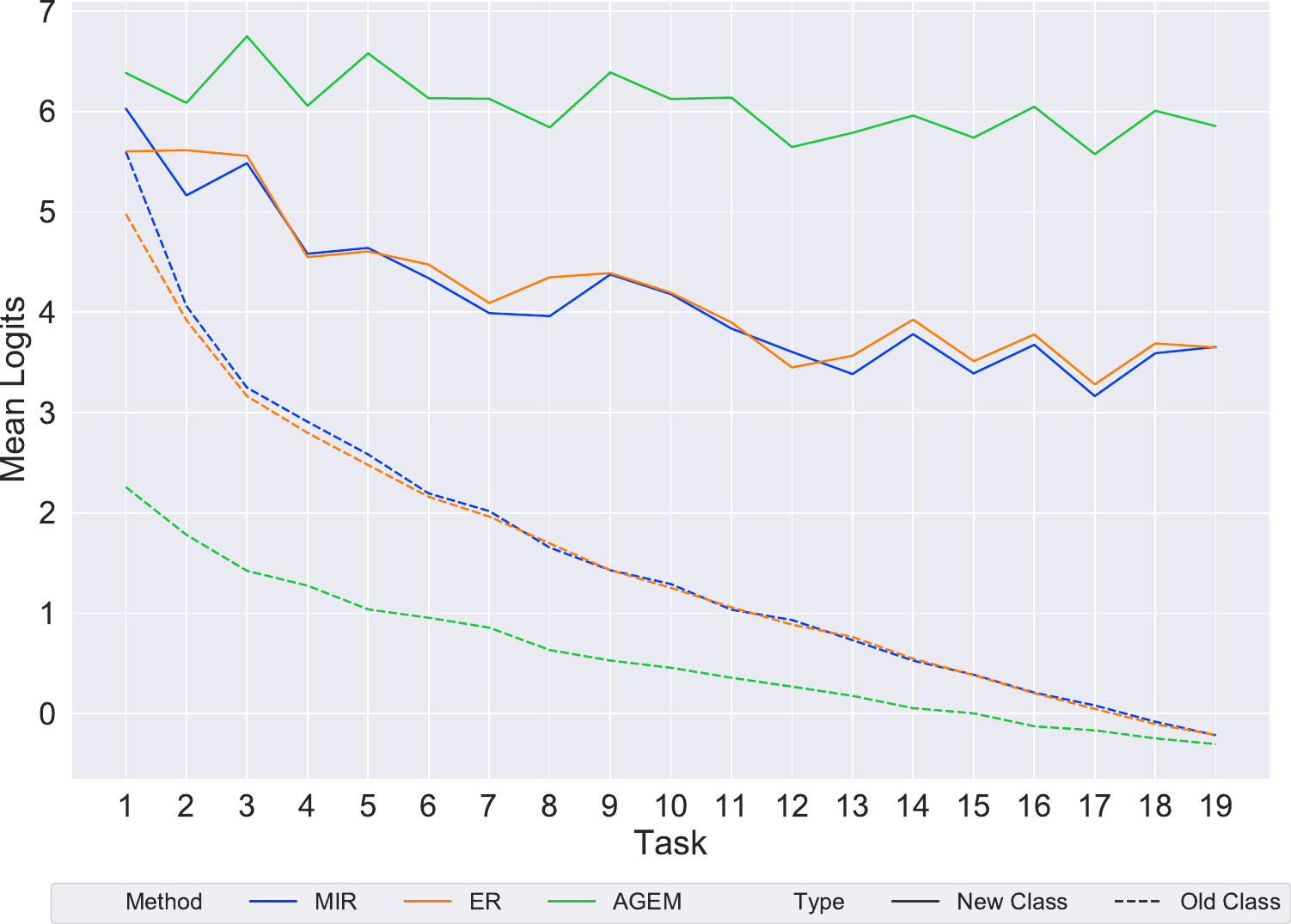}
    \caption{The mean logits for new and old classes}
    \label{fig:trick_motivation_2}
    \end{subfigure}
    \hfill
    \newline
    \newline
    \begin{subfigure}[t]{0.235\textwidth}
    \centering
    \includegraphics[height=3.1cm
    ]{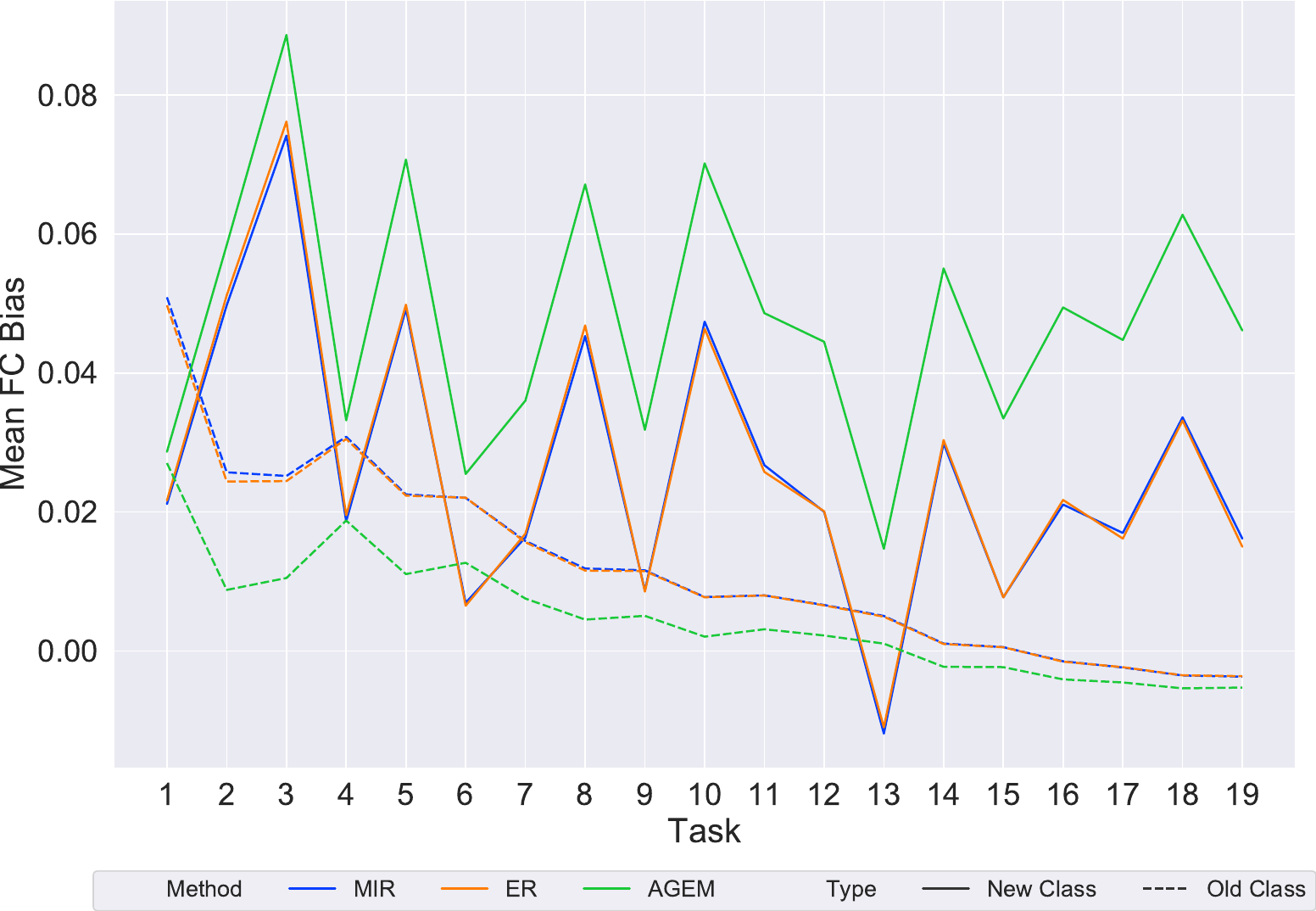}
    \caption{The mean of the bias terms in the last FC layer for new and old classes }
    \label{fig:trick_motivation_3}
    \end{subfigure}
    \hfill
    \begin{subfigure}[t]{0.235\textwidth}
    \centering
    \includegraphics[
    height=3.1cm
    ]{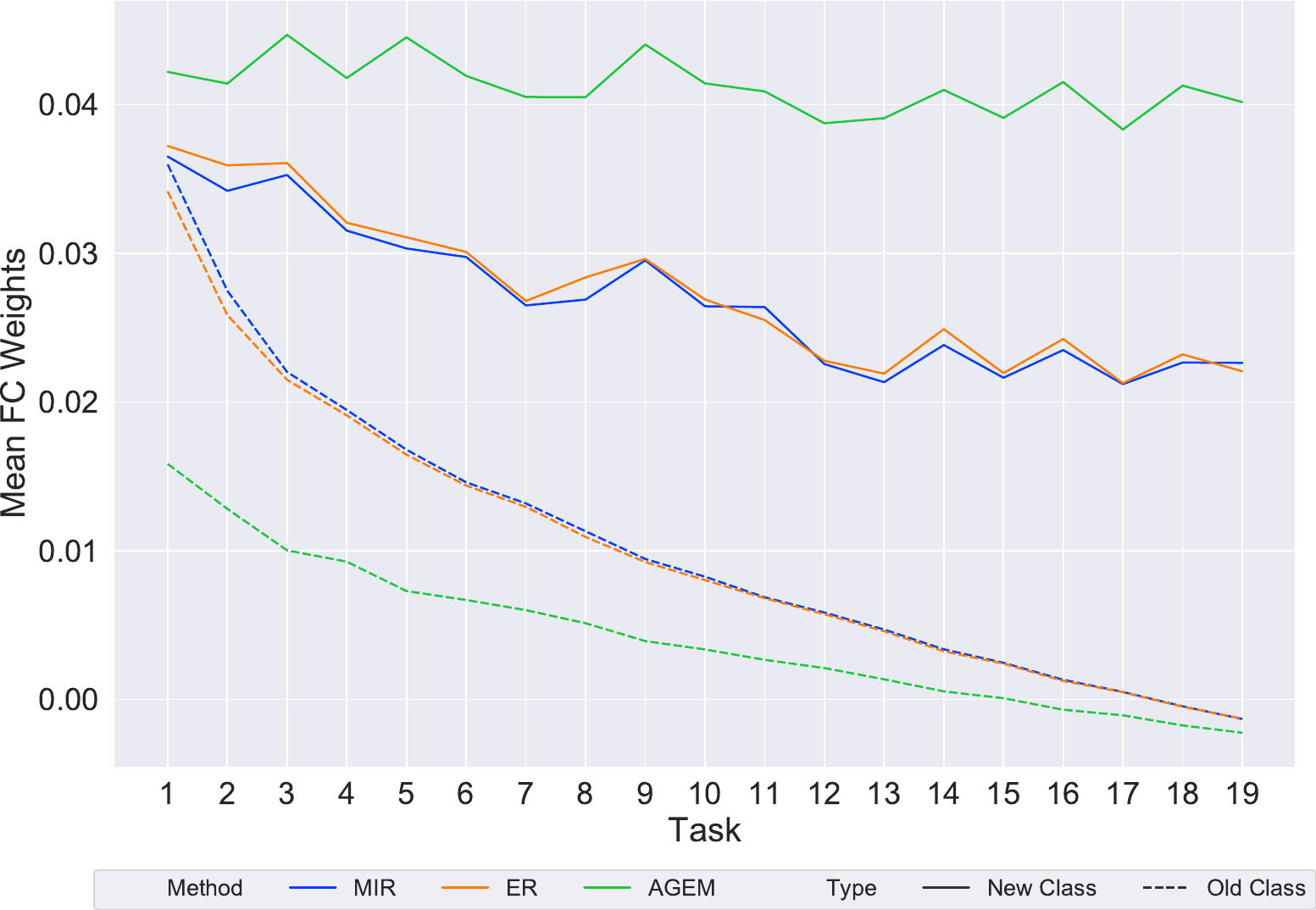}
    \caption{The mean of the weights in the last FC layer for new and old classes}
    \label{fig:trick_motivation_4}
    \end{subfigure}
    \caption{Error analysis for three memory-based methods(A-GEM, ER and MIR) with 5k memory buffer on Split CIFAR-100. The Error Ratio er(n,o) in (a) is the ratio of new class test samples misclassified as old classes to the total number of new class test samples. Same notation rule is applied to er(o, n).}\label{fig:trick_motivation}
\end{figure}

As shown in Table~\ref{table:trick_motivation}, all methods have strong bias towards new classes by the end of the training: A-GEM classifies all old class samples as new classes; ER and MIR misclassify 72\% and 66\% old class samples as new classes, respectively. Moreover, as we can see in Fig.~\ref{fig:trick_motivation_1}, $er(o, n)$ is higher than $er(n, o)$ most of the times along the training process for all three methods. These phenomena are not specific to the OCI setting, and \cite{SS, wa} also found similar results in the offline class incremental learning. Additionally, we easily find that ER and MIR are always better than A-GEM in terms of $er(o, n)$. This is because ER and MIR use the memory samples more directly, namely replaying them with the new class samples. The indirect use of memory samples in A-GEM is less effective in the class incremental setting.

To empirically verify the claim discussed in Section~\ref{biased_fc}, we also analyze the mean of logits for new and old classes. As shown in Fig.~\ref{fig:trick_motivation_2}, the mean of logits for new classes is always much higher than that for old classes, which explains the high $er(o, n)$ for all three methods. As we can see from Eq.~\eqref{eq: logit}, both feature extractor $\phi(\mathbf{x})$ and FC layer $\mathbf{W}$ may potentially contribute to the logit bias. However, previous works have found that even a small memory buffer (implying high class imbalance) can greatly alleviate catastrophic forgetting in the multi-head setting where the model can utilize the task-id to select the corresponding FC layer for each task~\cite{tiny, agem}. This suggests that the feature extractor is not heavily affected by the class imbalance, and therefore we hypothesize that the FC layer is biased. 

To validate the hypothesis, we plot the means of bias terms and weights in the FC layer for new and old classes in Fig.~\ref{fig:trick_motivation_3} and \ref{fig:trick_motivation_4}. As we can see, the means of weights for the new classes are much higher than those for the old classes, and the means of bias terms for the new classes are also higher than those for the old classes most of the times. Since the biased weights and bias terms in $\mathbf{W}$ have direct impacts on the output logits, the model has a higher chance of predicting a sample as new classes. \cite{SS, wu2019large, wa} have also verified the same hypothesis but with different validation methods. 

\begin{figure*}
    \centering
    \includegraphics[
    height=4.8cm]{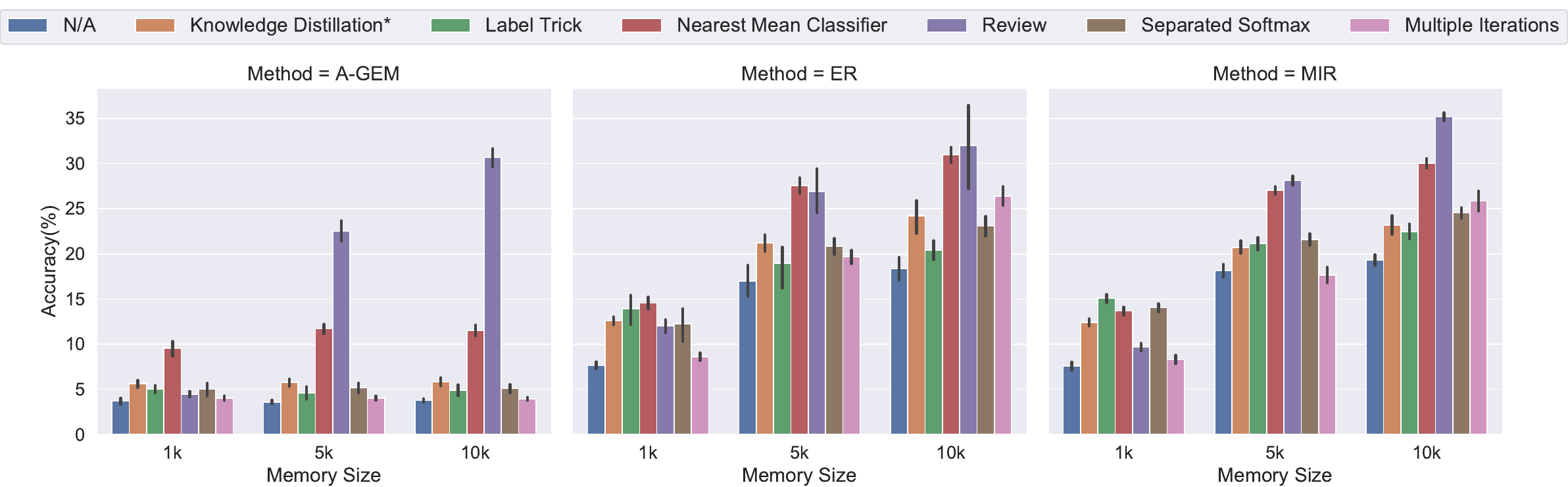}
    \caption{Comparison of various tricks for the OCI setting on Split CIFAR-100. We report average accuracy (end of training) for memory buffer with size 1k, 5k and 10k. N/A denotes the performance of the base methods (A-GEM, ER, MIR) without any trick.}\label{fig:trick_cifar}
\end{figure*}

\subsection{Effectiveness of tricks in the OCI setting}

\label{performance_tricks}
\begin{table*}[t!]
\tiny
\resizebox{\textwidth}{!}{
\begin{tabular}{l|ccc|ccc|ccc}
\toprule
Finetune & \multicolumn{9}{c}{$3.7\pm0.3$}\\
Offline & \multicolumn{9}{c}{$49.7\pm2.6$}\\
\midrule
Method& \multicolumn{3}{c|}{A-GEM}&\multicolumn{3}{c|}{ER}&\multicolumn{3}{c}{MIR}\\
\midrule
Buffer Size &M=1k& M=5k&M=10k&M=1k& M=5k&M=10k&M=1k&M=5k&M=10k\\
\midrule
    N/A & $3.7\pm0.4$ & $3.6\pm0.2$ & $3.8\pm0.2$&$7.6\pm0.5$ & $17.0\pm1.9$ & $18.4\pm1.4$&$7.6\pm0.5$ & $18.2\pm0.8$ & $19.3\pm0.7$\\
    LB   & $5.0\pm0.5$ & $4.6\pm0.8$ & $4.9\pm0.7$&$14.0\pm2.0$ & $19.0\pm2.6$ & $20.4\pm1.2$&$\mathbf{15.1\pm0.6}$ & $21.1\pm0.8$ & $22.5\pm0.9$\\
    KDC   & $8.3\pm0.7$ & $8.8\pm0.7$ & $7.7\pm1.1$&$11.7\pm0.8$ & $10.9\pm1.9$ & $11.9\pm2.2$&$12.0\pm0.5$ & $12.3\pm0.7$ & $11.8\pm0.6$\\
    KDC*  &  $5.6\pm0.5$ & $5.8\pm0.5$ & $5.8\pm0.5$&$12.6\pm0.5$ & $21.2\pm1.1$ & $24.2\pm1.9$&$12.4\pm0.5$ & $20.7\pm0.8$ & $23.2\pm1.2$\\
    MI &$4.0\pm0.3$&$4.0\pm0.3$&$4.0\pm0.2$&$8.6\pm0.5$&$19.7\pm0.9$&$26.4\pm1.2$&$8.5\pm0.4$&$17.7\pm1.0$&$25.9\pm1.2$\\
    SS   & $5.0\pm0.8$ & $5.2\pm0.6$ & $5.1\pm0.5$ &$12.3\pm2.1$ & $20.9\pm1.0$ & $23.1\pm1.2$&$14.0\pm0.5$ & $21.6\pm0.7$ & $24.5\pm0.7$\\
    NCM &   $\mathbf{9.5\pm0.9}$ & $11.7\pm0.6$ & $11.5\pm0.7$&$\mathbf{14.6\pm0.7}$ & $\mathbf{27.6\pm1.0}$ & $31.0\pm1.0$&$13.7\pm0.5$ & $27.0\pm0.5$ & $30.0\pm0.6$\\
    RV & $4.5\pm0.4$&$\mathbf{22.5\pm1.3}$&$\mathbf{30.7\pm1.2}$&$12.0\pm0.8$&$26.9\pm2.8$&$\mathbf{32.0\pm5.3}$&$9.7\pm0.5$&$\mathbf{28.1\pm0.6}$&$\mathbf{35.2\pm0.5}$\\

\midrule
Best OCI & $16.7\pm0.8$&${22.1\pm0.9}$&${28.8\pm0.9}$& $16.7\pm0.8$&${22.1\pm0.9}$&${28.8\pm0.9}$& $16.7\pm0.8$&${22.1\pm0.9}$&${28.8\pm0.9}$\\
\bottomrule
\end{tabular}}
\caption{Performance of compared tricks for the OCI setting on Split CIFAR-100. We report average accuracy (end of training) for memory buffer with size 1k, 5k and 10k. Best OCI refers to the best performance achieved by the compared methods in Table~\ref{table:nc}. }
\label{table:trick_cifar}
\end{table*}

\begin{table}
\centering
\scriptsize
\begin{tabular}{c c c c} 
    \toprule
    Trick& \multicolumn{3}{c}{Running Time(s)}\\
    \midrule
    &M=1k&M=5k&M=10k\\
    \cmidrule{2-4}
    N/A&83&82&84\\
    LB&87&88&89\\
    KDC&105&106&106\\
    KDC*&105&105&107\\
    MI&328&324&325\\
    SS&89&90&90\\
    NCM&126&282&450\\
    RV&98&159&230\\
    \bottomrule
\end{tabular}
\caption{Running time of different tricks applying to ER with different memory sizes on Split CIFAR-100. }
\label{tab:trick_time}
\end{table}

\begin{table*}
\resizebox{\textwidth}{!}{
\begin{tabular}{l|ccc|ccc|ccc|ccc}
\toprule
Method&\multicolumn{3}{c|}{Mini-ImageNet-Noise }&\multicolumn{3}{c|}{Mini-ImageNet-Occlusion}& \multicolumn{3}{c|}{Mini-ImageNet-Blur}&\multicolumn{3}{c}{CORe50-NI}\\
\midrule
Finetune &\multicolumn{3}{c|}{$11.1\pm1.0$}&\multicolumn{3}{c|}{$13.8\pm1.6$}& \multicolumn{3}{c|}{$2.4\pm0.2$}&\multicolumn{3}{c}{$14.0\pm2.8$}\\
Offline&\multicolumn{3}{c|}{$37.3\pm0.8$}&\multicolumn{3}{c|}{$38.6\pm4.7$} & \multicolumn{3}{c|}{$11.9\pm1.0$}&\multicolumn{3}{c}{$51.7\pm1.8$}\\
EWC&\multicolumn{3}{c|}{$12.5\pm0.8$}&\multicolumn{3}{c|}{$14.8\pm1.1$} & \multicolumn{3}{c|}{$2.6\pm0.2$}&\multicolumn{3}{c}{$11.6\pm1.5$}\\
LwF &\multicolumn{3}{c|}{$9.2\pm0.9$}&\multicolumn{3}{c|}{$12.8\pm0.8$}& \multicolumn{3}{c|}{$3.4\pm0.4$}&\multicolumn{3}{c}{$11.1\pm1.1$}\\
\midrule
Buffer Size &M=1k& M=5k&M=10k&M=1k& M=5k&M=10k&M=1k&M=5k&M=10k&M=1k&M=5k&M=10k\\
\midrule
ER &$\mathbf{19.4\pm1.3}$&$21.6\pm1.1$&$24.3\pm1.2$&$\mathbf{19.2\pm1.5}$&$\mathbf{23.4\pm1.4}$&$23.7\pm1.1$&$5.3\pm0.6$&$\mathbf{8.6\pm0.8}$&$9.4\pm0.7$&$24.1\pm4.2$&$28.3\pm3.5$&$30.0\pm2.8$\\
MIR &$18.1\pm1.1$&$\mathbf{22.5\pm1.4}$&$\mathbf{24.4\pm0.9}$&$17.6\pm0.7$&$22.0\pm1.1$&$\mathbf{23.8\pm1.2}$&$\mathbf{5.5\pm0.5}$&$8.1\pm0.6$&$9.6\pm1.0$&$\mathbf{26.5\pm1.0}$&$\mathbf{34.0\pm1.0}$&$\mathbf{33.3\pm1.7}$\\
GSS&$18.9\pm0.8$&$21.4\pm0.9$&$23.2\pm1.1$&$17.7\pm0.8$&$21.0\pm2.2$&$23.2\pm1.4$ &$5.2\pm0.5$&$7.6\pm0.6$&$8.0\pm0.6$&$25.5\pm2.1$&$27.2\pm2.0$&$25.3\pm2.1$\\
A-GEM &$14.0\pm1.3$&$14.6\pm0.7$&$14.2\pm1.4$&$16.4\pm0.7$&$13.9\pm2.6$&$14.4\pm2.0$&$4.4\pm0.4$&$4.4\pm0.4$&$4.3\pm0.5$&$12.4\pm1.1$&$13.8\pm1.2$&$15.0\pm2.2$\\
CN-DPM &$4.6\pm0.5$&-&-&$3.9\pm0.8$&-&-&$2.2\pm0.2$&-&-&$9.6\pm3.9$&-&-\\
GDumb &$5.4\pm1.0$&$12.5\pm0.7$&$15.2\pm0.5$&$5.4\pm0.4$&$14.2\pm0.6$&$20.2\pm0.4$&$3.3\pm0.2$&$7.5\pm0.2$&$\mathbf{10.0\pm0.2}$&$9.6\pm1.5$&$11.2\pm2.0$&$11.5\pm1.7$\\
\midrule
\bottomrule
\end{tabular}}
\caption{The Average Accuracy (end of training) for the ODI setting of Mini-ImageNet with three nonstationary types (Noise, Occlusion, Blur) and CORe50-NI.}
\label{table:ni}
\end{table*}
We evaluate Label trick (LB), Knowledge Distillation and Classification (KDC), Multiple Iterations (MI), Separated Softmax(SS), Nearest Class Mean (NCM) classifier, Review trick(RV) on three memory-based methods: A-GEM, ER and MIR. The results are shown in Table~\ref{table:trick_cifar} and Fig.~\ref{fig:trick_cifar}. 

Firstly, although all tricks enhance the basic A-GEM, only RV can bring A-GEM closer to ER and MIR, which reiterates that the direct replay of the memory samples is more effective than the gradient projection approach in A-GEM.

For replay-based ER and MIR, LB and NCM are the most effective when M=1k and can improve the accuracy by around 100\% (7.6\% $\rightarrow$ 14.5\% on average). KDC, KDC* and  SS have similar performance improvement effect and can boost the accuracy by around 64\% (7.6\% $\rightarrow$ 12.4\% on average).  With a larger memory size, NCM remains very effective, and RV becomes much more helpful. When M=10k, RV boosts ER's performance by 74\% (18.4\% $\rightarrow$ 32.0\%) and improve MIR's performance by 82\% (from 19.3\% $\rightarrow$ 35.2\%). Also, KDC fails with a larger memory due to over-regularization of the KD term, and the modified KDC* has much better performance. Compared with other tricks, MI and RV are more sensitive to the memory size since these tricks highly depend on the memory. Note that when equipped with NCM or RV, both ER and MIR can outperform the best performance achieved by the compared methods (Table~\ref{table:nc}) when M=5k or 10k. 

As shown in Table~\ref{tab:trick_time}, the running times of LB, KDC, KDC*, MI and SS do not depend on the memory size and have a limited increase compared to the baseline. Since NCM needs to calculate the means of all the classes in the memory and RV requires additional training of the whole memory before evaluation, their running times increase as the memory size grows. 

To sum up, all of the tricks are beneficial to the base methods (A-GEM, ER, MIR); NCM is a useful and robust trick across all memory sizes, while LB and RV are more advantageous in smaller and larger memory, respectively. The running times of NCM and RV go up with the increase in the memory size, and other tricks only add a fixed overhead to the running time. We also get similar results in Split Mini-ImageNet, as shown in~\ref{additional}.

\subsection{Performance comparison in Online Domain Incremental (ODI) setting}
\label{perf_domain_incremental}

\begin{figure*}
    \centering
    \includegraphics[
    height=13cm]{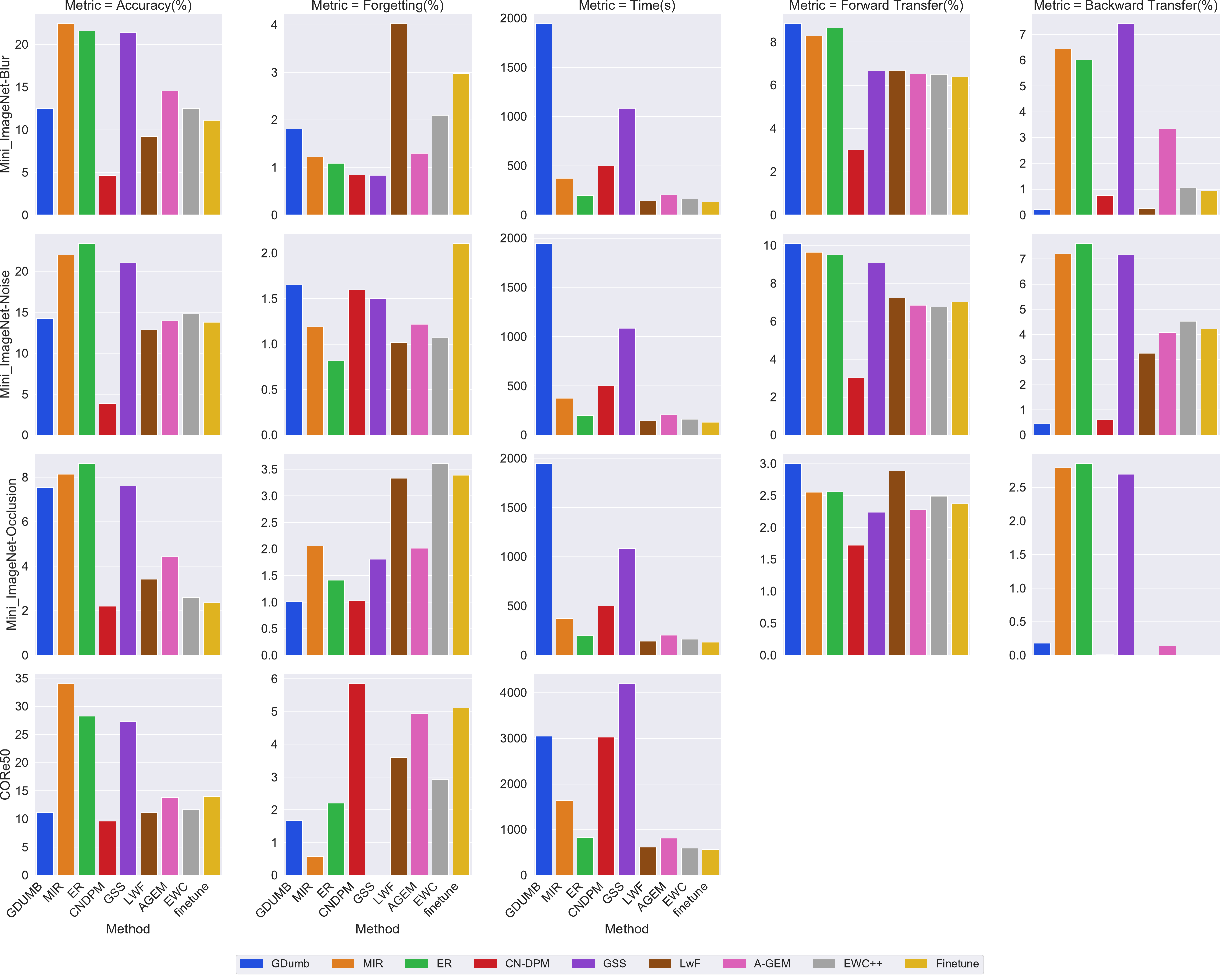}
    \caption{Average Accuracy, Forgetting, Running Time, Forward Transfer and Backward Transfer for the ODI setting with a 5k memory buffer. Each column represents a metric and each row represents a dataset. Forward and backward transfer are not applicable in CORe50 since it uses one test set for all tasks. }    
    \label{fig:ni_bar}
\end{figure*}

Since most of the surveyed methods are only evaluated in the class incremental setting in the original papers, we evaluate them in the ODI setting to investigate their robustness and ability to generalize to other CL settings. We assess the methods with CORe50-NI---a dataset designed for ODI---and the proposed NS-MiniImageNet dataset consisting of three nonstationary types: noise, occlusion, and blur (see Table~\ref{tab: ns_mini}). The average accuracy at the end of training is summarized in Table~\ref{table:ni}. 

Generally speaking, all replay-based methods (ER, MIR, GSS) show comparable performance across three memory sizes and outperform all other methods. GDumb, the strong baseline that dominates the OCI setting in most cases, is no longer as competitive as the replay-based methods and fails completely in CORe50-NI. One of the reasons is that class imbalance, the key cause of forgetting in the OCI setting, does not exist in ODI since the class labels are the same for all tasks. Moreover, in the ODI setting, samples in the data stream change gradually and smoothly with different nonstationary strengths (NS-MiniIMageNet) or nonstationary types (CORe50-NI). Learning new samples sequentially with the replay samples(replay-based) may be more effective for the model to adapt to the gradually changing nonstationarity than learning only the samples in the buffer (GDumb). Additionally, the greedy memory update strategy (see Algorithm~\ref{alg:gdumb} in Appendix) in GDumb is not suitable for the ODI setting as the buffer will comprise mostly of samples in the latest tasks due to the greedy update, and GDumb will have very limited access to samples in the earlier tasks. Using reservoir sampling as the update strategy may alleviate this shortcoming since reservoir sampling ensures every data point has the same probability to be stored in the memory.

CN-DPM has terrible performance in this setting because it is very sensitive to hyperparameters, and the method cannot find the hyperparameter set that works in this setting within the same tuning budget as other methods. 

Regarding methods without a memory buffer, the KD-based LwF underperforms EWC and Finetune, implying KD may not be useful in the ODI setting. 

Another interesting observation is that all methods, including Offline, show unsatisfactory results in the blur scenario. The pivotal cause may be due to the backbone model we use (ResNet18~\cite{he2016deep}) since a recent study points out that Gaussian blur can easily degrade the performance of ResNet~\cite{roy2018effects}. 

In terms of other metrics, as shown in Fig.~\ref{fig:ni_bar}, we find that all methods show positive forward transfer, and replay-based methods show much better backward transfer than others. The first reason is that tasks in the ODI setting share more cross-task resemblances than the OCI setting; secondly, the bias towards the current task due to class imbalance does not happen since new tasks contain the same class labels as old tasks. Thus, the model is able to perform zero-shot learning (forward transfer) and improve the preceding tasks (backward transfer). 

In summary, replay-based methods exhibit more robust and surpassing performance in the ODI setting. Considering the running time and the performance in the larger scale dataset, MIR stands out as a versatile and competitive method in this setting.

\begin{table}
\centering
\scriptsize
\begin{tabular}{m{1cm} |m{7cm}} 
    \toprule
    Method&Comments\\
    \midrule
    \multicolumn{2}{l}{\textbf{Regularization-based}}\\
        \midrule
    EWC++&
\begin{itemize}
\item Ineffective in both OCI and ODI settings
\item Suffers from gradient explosion
\end{itemize} \\
    LwF& \begin{itemize}
        \item Effective on small scale datasets in OCI (achieves similar performance as replay-based methods with small memory)
        \item Ineffective in ODI setting
    \end{itemize}\\
        \midrule
    \multicolumn{2}{l}{\textbf{Memory-based}}\\
        \midrule
    ER&\begin{itemize} 
        \item Efficient training time over other memory-based methods
        \item Better than GSS in most cases but worse than MIR, especially with a large memory buffer
    \end{itemize}\\
    MIR&\begin{itemize}
        \item A versatile and competitive method in both OCI and ODI settings
        \item Works better on a large scale dataset and a large memory buffer
    \end{itemize}\\
    GSS & \begin{itemize}
        \item Inefficient training time
        \item Worse than other memory-based methods in most cases
    \end{itemize}\\
    iCaRL&\begin{itemize}
        \item Best performance (with large margins) with a small memory buffer on small scale datasets 
    \end{itemize}\\
    A-GEM&\begin{itemize}
        \item Ineffective in both OCI and ODI settings
    \end{itemize}\\
    GDumb&\begin{itemize}
        \item Best performance with a large memory buffer on small scale datasets in OCI setting
        \item Ineffective in ODI mostly due to its memory update strategy
        \item Inefficient training time due to training from scratch at every inference point

    \end{itemize}\\
     \midrule
    \multicolumn{2}{l}{\textbf{Parameter-isolation-based}}\\
     \midrule
    CN-DPM&\begin{itemize}
        \item Effective when memory size is small
        \item Sensitive to hyperparameters and when testing on a new dataset, it may not  find a working hyperparameter set given the same tuning budget as others
        \item Longest training time among compared methods
    \end{itemize}\\
\bottomrule
\end{tabular}
\caption{Overall comments for compared methods}
\label{tab: comment_method}
\end{table}


\subsection{Overall comments for methods and tricks}
\label{comment}
We summarize the key findings and give comments on each method and trick based on our findings in Table~\ref{tab: comment_method} and Table~\ref{tab:comment_trick}.

\section{Trendy Directions in Online CL}
\label{future}
In this section, we discuss some emerging directions in online CL that have attracted interest and are expected to gain more attention in the future.
\paragraph{Raw-Data-Free Methods} 
In some applications, storing raw images is not feasible due to privacy and security concerns, and this calls for CL methods that maintain reasonable performance without storing raw data. 
Regularization~\cite{lwf,Kirkpatrick2017} is one of the directions but \cite{lesort2019regularization} shows that this approach has theoretical limitations in the class incremental setting and cannot be used alone to reach decent performance. We also have empirically confirmed their claims in this work. Generative replay~\cite{wu2018memory, generative} is another direction but it is not viable for more complex datasets as the current deep generative models still cannot generate satisfactory images for such datasets~\cite{mir, lesort2019generative}. 

Feature replay is a promising direction where latent features of the old samples at a given layer (feature extraction layer) are relayed instead of raw data~\cite{iscen2020memory, hayes2020remind, liu2020generative, pellegrini2019latent, wang2021acae}. Since the model changes along the training process, to keep the latent features valid, \cite{pellegrini2019latent} proposes to slow-down---in the limit case, freeze---the learning of all the layers before the feature extraction layer, while \cite{iscen2020memory} proposes a feature adaptation method to map previous features to their correct values as the model is updated. Another way is to generate latent features with a deep generative model~\cite{liu2020generative}. 

There are other lately proposed approaches which do not require storing the raw data~\cite{madireddy2020neuromodulated, yu2020semantic, chen2020mitigating, chopra2005learning, dmc, hayes2020lifelong,fini2020online}. For example, SDC~\cite{yu2020semantic} leverages embedding networks~\cite{chopra2005learning} and the nearest class mean classifier~\cite{ncm}. The approach proposes a method to estimate the drift of features during learning the current task and compensate for the drift in the absence of previous samples. DMC~\cite{dmc} trains a separate model for new tasks and combines the new and old models using publicly available unlabeled data via a double distillation training objective. DSLDA~\cite{hayes2020lifelong} freezes the feature extractor and uses deep Streaming Linear Discriminant Analysis~\cite{pang2005incremental} to train the output layer incrementally. InstAParam~\cite{chen2020mitigating} leverages the "instance awareness" where each data instance is classified
by a path in the network searched by the controller from a meta-graph (updated online). It encourages or restricts the gradient updates for the current path based on the similarity between the current instance and previous instances. EBM-CL\cite{li2020energy} interprets continual learning for classification as learning an Energy-Based Model~\cite{lecun2006tutorial} across classes and proposes a contrastive divergence based training objective, which can naturally handle dynamically growing numbers
of classes. 

With the increasing data privacy and security concerns, the raw-data-free methods are expected to attract more research endeavour in the coming years. 

\paragraph{Meta Learning}
Meta-learning is an emerging learning paradigm where a neural network evolves from multiple related learning episodes and generalizes the learned knowledge to unseen tasks~\cite{metasurvey}. Since meta-learning builds up a potential framework to advance CL, a lot of meta-learning based CL methods have been proposed recently, and most of them support the online setting. MER~\cite{mer} combines experience replay with optimization based meta-learning to maximize transfer and minimize interference based on future gradients. OML~\cite{javed2019meta} is a meta-objective that uses interference as a training signal to learn a representation that accelerates future learning and avoid catastrophic interference. More recently,   iTAML~\cite{rajasegaran2020itaml} proposes to learn a task-agnostic model that automatically predicts the task and quickly adapts to the predicted task with meta-update. La-MAML~\cite{gupta2020maml} proposes an efficient gradient-based meta-learning that incorporates per-parameter learning rates for online CL. MERLIN~\cite{merlin} proposes an online CL method based on consolidation in a meta-space, namely, the latent space that generates model weights for solving downstream tasks. In~\cite{caccia2020online}, authors propose Continual-MAML, an online extension of MAML~\cite{maml}, that can cope the new CL scenario they propose. We believe meta-learning based online CL methods will continue to be popular with recent advances in meta-learning. 


\paragraph{CL in Other Areas}
Although image classification and reinforcement learning are the main focuses for most CL works, CL has drawn more and more attention in other areas. Object detection has been another emerging topic in CL, and multiple works have been proposed lately to tackle this problem. Most methods leverage KD~\cite{distill} to alleviate CF, and the main differences between the methods are the base object detector and distillation parts in the network~\cite{shmelkov2017incremental, od2, od3, liumulti}. More recently, a meta-learning based approach is proposed to reshape model gradients for better information share across incremental tasks~\cite{joseph2020incremental}. A replay-based method is introduced to address streaming object detection by replaying compressed representation in a fixed memory buffer~\cite{acharya2020rodeo}. 

Beyond computer vision, CL with sequential data and recurrent neural network (RNN) has gained attention over the past few years. Recent works have confirmed that RNNs, including LSTMs, are also immensely affected by CF~\cite{sodhani2020toward, schak2019study, arora2019does}. In \cite{sodhani2020toward}, the authors unify GEM~\cite{gem} and Net2Net~\cite{chen2015net2net} to tackle forgetting in RNN. More recently, \cite{ehret2020continual} shows that weight-importance based CL in RNNs are limited and that the hypernetwork-based approaches are more effective in alleviating forgetting. Meanwhile, \cite{duncker2020organizing} proposes a learning rule to preserve network dynamics within subspaces for previous tasks and encourage interfering dynamics to explore orthogonal subspaces when learning new tasks. Moreover, multiple works are proposed to address general CL language learning~\cite{sun2019lamol, li2019compositional} and specific language tasks, such as dialogue systems~\cite{mi2020continual, mazumder2019lifelong, liu2020lifelong}, image captioning~\cite{del2020ratt}, sentiment classification~\cite{ke2020continual} and sentence representation learning~\cite{liu2019continual}. 

Recommender systems have also started to adopt CL~\cite{mi2020memory, yuan2020one, mi2020ader, xu2020graphsail}. ADER~\cite{mi2020ader} is proposed to handle CF in session-based recommendation using the adaptive distillation loss and replay with heading~\cite{welling2009herding} technique. GraphSAIL~\cite{xu2020graphsail} is introduced for Graph Neural Networks based recommender systems to preserve a user's long-term preference during incremental model updates using local structure distillation, global structure distillation and self-embedding distillation. 

Several works also address the deployment of CL in practice. \cite{liu2020learning} introduces on-the-job learning, which requires a deployed model to discover new tasks, collect training data continuously and incrementally learn new tasks without interrupting the application. The author also uses chat-bots and self-driving cars as the examples to highlight the necessity of on-the-job learning. \cite{diethe2019continual} presents a reference architecture for self-maintaining intelligent systems that can adapt to shifting data distributions, cope with outliers, retrain when necessary, and learn new tasks incrementally. \cite{lee2020clinical} discusses the clinical application of CL from three perspectives: diagnosis, prediction and treatment decisions. \cite{lange2020unsupervised} addresses a practical scenario where a high-capacity server interacts with a large group of resource-limited edge devices and proposes a Dual User-Adaptation framework which disentangles user-adaptation into model personalization on the server and local data regularization on the user device. 
 
\begin{table}
\scriptsize
\centering
\begin{tabular}{m{.7cm} |m{7cm}} 
    \toprule
    Trick&Comments\\
    \midrule
    LB&
\begin{itemize}
\item Effective when memory buffer is small
\item Fixed and limited training time overhead
\end{itemize} \\
\midrule
KDC&\begin{itemize}
    \item Fails because of over-regularization of knowledge distillation loss
\end{itemize}\\
\midrule
KDC*&\begin{itemize}
    \item Provides moderate improvement with fixed and acceptable training time overhead
\end{itemize}\\
\midrule
MI&\begin{itemize}
    \item Better improvement with a larger memory buffer
    \item Training time increases with more iterations
\end{itemize}\\
\midrule
SS&\begin{itemize}
    \item Similar improvement as KDC* but with less training time overhead
\end{itemize}\\
\midrule
NCM&\begin{itemize}
    \item Provides very strong improvement across different memory sizes.
    \item Baselines equipped with it outperform state-of-the-art methods when the memory buffer is large
    \item Inference time increases with the growth of the memory size
\end{itemize}\\
\midrule
RV&\begin{itemize}
    \item Presents very competitive improvement, especially with a larger memory buffer
    \item Baselines equipped with it outperform state-of-the-art methods
    \item Training time increases with the growth of the memory size but it is more efficient than NCM
\end{itemize}\\
\bottomrule
\end{tabular}
\caption{Overall comments for compared tricks}
\label{tab:comment_trick}
\end{table}

\section{Conclusion}
\label{conclusion}
To better understand the relative advantages of recently proposed online CL approaches and the settings where they work best, we performed extensive experiments with nine methods and seven tricks in the online class incremental (OCI) and online domain incremental (ODI) settings. 

Regarding the performance in the OCI setting (see Table~\ref{table:nc}, Fig.~\ref{fig:nc_5k} and~\ref{fig:nc_bar}), we conclude:
\begin{itemize}
    \item For memory-free methods, LwF is effective in CIFAR100 and Mini-ImageNet, showing similar performance as replay-based methods with a small memory buffer. However, all memory-free methods fail in the larger CORe50-NC. 
    \item When the memory buffer is small, iCaRL shows the best performance in CIFAR100 (a small margin) and Mini-ImageNet (a large margin), followed by CN-DPM. 
    \item With a larger memory buffer, GDumb---a simple baseline---outperforms methods designed specifically for the CL problem in CIFAR100 and Mini-ImageNet at the expense of much longer training times.
    \item In the larger and more realistic CORe50-NC dataset, MIR consistently surpasses all the other methods across different memory sizes.  
    \item We experimentally and theoretically confirm that a key cause of CF is the bias towards new classes in the last fully connected layer due to the imbalance between previous data and new data~\cite{wu2019large, wa, SS}.
    \item None of the methods show any positive forward and backward transfer due to the bias mentioned above. 
\end{itemize}

The conclusions from our experiments for the OCI tricks (see Table~\ref{table:trick_cifar}, Fig.~\ref{fig:trick_cifar}) are as follows:
\begin{itemize}
    \item When the memory size is small, LB and NCM are the most effective, showing around 64\% relative improvement.
    \item With a larger memory buffer, NCM remains effective, and RV becomes more helpful, showing around 80\% relative improvement.
    \item When equipped with NCM or RV, both ER and MIR can outperform the best performance of the compared methods without tricks. 
    \item The running times of NCM and RV increase with the growth in memory size, but other tricks only add a fixed overhead to the running time.
\end{itemize}

For the ODI setting (see Table~\ref{table:ni}, Fig.~\ref{fig:ni_bar}), we conclude:
\begin{itemize}
    \item Generally speaking, all replay-based methods (ER, MIR, GSS) show comparable performance across three memory sizes and outperform all other methods.
    \item GDumb, the strong baseline that dominates the OCI setting in most cases, is no longer effective, possibly due to its memory update strategy.
    \item Other OCI methods cannot generalize to the ODI setting. 
\end{itemize}

Detailed comments for compared methods and tricks can be found in Table~\ref{tab: comment_method} and Table~\ref{tab:comment_trick}. 

In conclusion, by leveraging the best methods and tricks identified in this comparative survey, 
online CL (with a very small mini-batch) is now approaching offline performance, bringing CL much closer to its ultimate goal of matching offline training that opens up CL for effective deployment on edge and other RAM-limited devices. 



\section{Acknowledgement}
This research was supported by LG AI Research.

\clearpage
\bibliographystyle{elsarticle-num} 
\bibliography{references}







\clearpage
\appendix
\section{Algorithms}
\label{app:algo}

In this section, we provide the algorithms for different memory update strategies described in Section~\ref{methods}. 
\setcounter{algocf}{0}
\renewcommand{\thealgocf}{A\arabic{algocf}}
\begin{algorithm}
\footnotesize
\caption{Reservoir sampling}
\label{alg:reservoir}
  \SetKwProg{myproc}{procedure}{ MemoryUpdate $(mem_sz,t,n,B)$}{}
  \myproc{}{
  \nl $j \leftarrow 0$\\
  \nl \For{$(\mathrm{x}, y)$ in $B$}{
  \nl $\quad M \leftarrow|\mathcal{M}|$ \Comment{Number of samples currently stored in the memory}\\
  \nl \eIf{$M<\mathrm{mem}_{-} \mathrm{sz}$}{
  \nl $\quad \mathcal{M}$.append $(\mathrm{x}, y, t)$}{
  \nl $i=\operatorname{randint}(0, n+j)$\\
  \nl  \uIf{$i<\mathrm{mem}_{-} \mathrm{sz}$}{
  \nl $\mathcal{M}[i] \leftarrow(\mathbf{x}, y, t)$ \Comment{Overwrite memory slot}}}
  \nl $\quad j \leftarrow j+1$\\
  }
\textbf{return} $\mathcal{M}$
  }
\end{algorithm}

\begin{algorithm}
\footnotesize
\caption{GSS-Greedy}
\label{alg:GSS}
\SetCustomAlgoRuledWidth{0.49\textwidth}
\SetAlgorithmName{Algorithm}{}{}
\SetKwInOut{Input}{Input~}
     \SetKwInput{Initialize}{Initialize}
     \SetKwInput{Receive}{Receive}
     \SetKwInput{Update}{Update}
\nl \Input{$n, M$}
\nl \Initialize{$\mathcal{M}, \mathcal{C}$}
\nl \Receive{$(x, y)$}
\nl \Update{$(x, y, \mathcal{M})$}
\nl $X, Y \leftarrow$ RandomSubset $(\mathcal{M}, \mathrm{n})$\\
\nl $g \leftarrow \nabla \ell_{\theta}(x, y) ; G \leftarrow \nabla_{\theta} \ell(X,Y)$\\
\nl $c=\max _{i}\left(\frac{\left\langle g, G_{i}\right\rangle}{\|g\|\left\|G_{i}\right\|}\right)+1$ \Comment{make the score positive}\\
\nl \If{$\operatorname{len}(\mathcal{M})>=M$}{
\nl \If{$c<1$}{\Comment{cosine similarity $<0$}\\
\nl $i \sim P(i)=\mathcal{C}_{i} / \sum_{j} \mathcal{C}_{j} $\\
\nl $r \sim \text { uniform }(0,1)$\\
\nl \If{$r<\mathcal{C}_{i} /\left(\mathcal{C}_{i}+c\right)$}{
\nl $\quad \mathcal{M}_{i} \leftarrow(x, y) ; \mathcal{C}_{i} \leftarrow c$}
\nl \textbf{end if}
}
\nl \textbf{end if}
}
\nl \Else{
\nl $\mathcal{M} \leftarrow \mathcal{M} \cup\{(x, y)\} ; \mathcal{C} \cup\{c\}$}
\nl \textbf{end if}

\end{algorithm}

\begin{algorithm}
\footnotesize
\caption{Greedy Balancing Sampler}
\label{alg:gdumb}
\SetCustomAlgoRuledWidth{0.49\textwidth}
\SetAlgorithmName{Algorithm}{}{}
\SetKwInOut{Input}{Input~}
     \SetKwInput{Init}{Init}
\nl \Init{counter $C_{0}=\{\}, \mathcal{D}_{0}=\{\}$ with capacity $k .$ Online samples arrive from $\mathrm{t}=1$}
\nl \textbf{function}{ SAMPLE $\left(x_{t}, y_{t}, \mathcal{D}_{t-1}, \mathcal{Y}_{t-1}\right) \quad$}\Comment{Input: New sample and past state}\\
\nl \quad $k_{c}=\frac{k}{\left|\mathcal{Y}_{t-1}\right|}$ \\
\nl \quad \If{$y_{t} \notin \mathcal{Y}_{t-1}$ or $C_{t-1}\left[y_{t}\right]<k_{c}$}{
\nl \quad \If {$\sum_{i} C_{i}>=k$}{\Comment{If memory is full, replace}\\
\nl \quad $y_{r}=\operatorname{argmax}\left(C_{t-1}\right)$\Comment{Select largest class, break ties randomly}\\
\nl \quad $\left(x_{i}, y_{i}\right)=\mathcal{D}_{t-1} \cdot \text { random }\left(y_{r}\right)$\Comment{Select random sample from class $y_{r}$}\\
\nl \quad $\mathcal{D}_{t}=\left(\mathcal{D}_{t-1}-\left(x_{i}, y_{i}\right)\right) \cup\left(x_{t}, y_{t}\right)$\\
\nl \quad $C_{t}\left[y_{r}\right]=C_{t-1}\left[y_{r}\right]-1$}
\nl \quad \Else{\Comment{If memory has space, add}\\
\nl \quad $\mathcal{D}_{t}=\mathcal{D}_{t-1} \cup\left(x_{t}, y_{t}\right)$}
\nl \quad \textbf{end if}\\
\nl \quad $\mathcal{Y}_{t}=\mathcal{Y}_{t-1} \cup y_{t}$\\
\nl \quad $C_{t}\left[y_{t}\right]=C_{t-1}\left[y_{t}\right]+1$}
\nl \quad \textbf{end if}\\
\nl \quad \textbf{return $\mathcal{D}_{t}$}\\
\nl \textbf{end function}\\
\end{algorithm}

\section{Experiment Details}
\label{app:exp_detail}
\subsection{Dataset Detail}

The summary of dataset statistics is provided in Table~\ref{app:dataset}. 

The strength of each nonstationary type used in the experiments are summarized below. 
\begin{itemize}
    \item Noise: [0.0, 0.4, 0.8, 1.2, 1.6, 2.0, 2.4, 2.8, 3.2, 3.6]
    \item Occlusion: [0.0, 0.07, 0.13, 0.2, 0.27, 0.33, 0.4, 0.47, 0.53, 0.6]
    \item Blur: [0.0, 0.28, 0.56, 0.83, 1.11, 1.39, 1.67, 1.94, 2.22, 2.5]
\end{itemize}
\setcounter{table}{0}
\begin{table*}[t!]
\centering
\scriptsize
\begin{tabular}{l | c c c c c c}
    
    \toprule
    Dataset &\#Task&\#Train/task&\#Test/task&\#Class&Image Size&Setting\Bstrut \\ 
    \hline\hline
    \Tstrut
    Split MiniImageNet& 20&2500&500&100&3x84x84&OCI\\
    Split CIFAR-100&20&2500&500&100&3x32x32&OCI\\
    CORe50-NC&9&12000$\sim$24000&4500$\sim$9000&50&3x128x128&OCI\\
    NS-MiniImageNet&10&5000&1000&100&3x84x84&ODI\\
    CORe50-NI&8&15000&44972&50&3x128x128&ODI\\
    \bottomrule
\end{tabular}
\caption{Summary of dataset statistics}
\label{app:dataset}
\end{table*}

\subsection{Implementation Details}
\label{app:impl_detail}
This section describes the implementation details of each method, including the hyperparameter grid considered for each dataset (see Table~\ref{table:hyper}). As we described in Section~\ref{hyperparameter} of the main paper, the first $D^{CV}$ tasks are used for hyperparameter tuning to satisfy the requirement that the model does not see the data of a task more than once, and $D^{CV}$ is set to 2 in this work. 

\begin{itemize}
    \item EWC++: We set the $\alpha$ in Eq.~\eqref{eq:ewc_online} to 0.9 as suggested in the original paper. We tune three hyperparameters in EWC++, learning rate (LR), weight decay(WD) and $\lambda$ in Eq.~\eqref{eq:ewc}.
    \item LwF: We set the temperature factor $T=2$ as the original paper and other CL papers. The coefficient $\lambda$ for $\mathcal{L}_{KD}$ is set to $\frac{\abs{C_{new}}}{\abs{C_{old}}+\abs{C_{new}}}$ following the idea from~\cite{wu2019large} and the coefficient for $\mathcal{L}_{CE}$ is set to $1-\lambda$. 
    \item ER: The reservoir sampling used in \update{} follows Algorithm~\ref{alg:reservoir} in~\ref{app:algo}. For \retrieval{}, we randomly select samples with mini-batch size of 10 irrespective of the size of the memory buffer. 
    \item MIR: To reduce the computational cost, MIR selects $C$ random samples from the memory buffer as the candidate set to perform the criterion search. We tune LR, WD as well as $C$. 
    \item GSS: For every incoming sample, GSS computes the cosine similarity of the new sample gradient to $n$ gradient vectors of samples randomly drawn from the memory buffer (see Algorithm~\ref{alg:GSS} in~\ref{app:algo}). Other than LR and WD, we also tune $n$. 
    \item iCaRL: We replace the herding-based~\cite{welling2009herding} memory update method with reservoir sampling to accommodate the online setting. We use random sampling for \retrieval{} and tune LR and WD.  
    \item A-GEM: We use reservoir sampling for \update{} and random sampling for \retrieval{} and tune LR and WD. 
    \item CN-DPM: CN-DPM is much more sensitive to hyperparameters than others. We need to use different hyperparameter grids for different scenarios and datasets. Other than LR, we tune $\alpha$, the concentration parameter controlling how sensitive the model is to new data and classifier\_chill $cc$, the parameter used to adjust the VAE loss to have a similar scale as the classifier loss. 
    \item GDumb: We use batch size of 16 and 30 epochs for all memory sizes. We clip gradient norm with max norm 10.0 and tune LR and WD.  
\end{itemize}
\begin{table*}[t!]
\resizebox{\textwidth}{!}{
\begin{tabular}{l|lllll}
\toprule
Method&CIFAR-100&Mini-ImageNet&CORe50-NC&NS-MiniImageNet&CORe50-NI\\
\midrule
\multirow{2}{*}{EWC++}&\multicolumn{5}{c}{LR: [0.0001, 0.001, 0.01, 0.1]}\\
&\multicolumn{5}{c}{WD: [0.0001, 0.001], $\lambda$: [0, 100, 1000]}\\\cline{2-6}
\multirow{2}{*}{LwF}&\multicolumn{5}{c}{LR: [0.0001, 0.0003, 0.001, 0.003, 0.01, 0.03, 0.1]}\\
&\multicolumn{5}{c}{WD: [0.0001, 0.001, 0.01, 0.1]}\\\cline{2-6}
\multirow{2}{*}{ER}&\multicolumn{5}{c}{LR: [0.0001, 0.0003, 0.001, 0.003, 0.01, 0.03, 0.1]}\\
&\multicolumn{5}{c}{WD: [0.0001, 0.001, 0.01, 0.1]}\\\cline{2-6}
\multirow{2}{*}{MIR}&\multicolumn{5}{c}{LR:[0.0001, 0.001, 0.01, 0.1]}\\
&\multicolumn{5}{c}{WD: [0.0001, 0.001], $C$: [25, 50, 100]}\\\cline{2-6}

\multirow{2}{*}{GSS}&\multicolumn{5}{c}{LR:[0.0001, 0.001, 0.01, 0.1]}\\
&\multicolumn{5}{c}{WD: [0.0001, 0.001], $n$: [10, 20, 50]}\\\cline{2-6}

\multirow{2}{*}{iCaRL}&\multicolumn{5}{c}{LR: [0.0001, 0.0003, 0.001, 0.003, 0.01, 0.03, 0.1]}\\
&\multicolumn{5}{c}{WD: [0.0001, 0.001, 0.01, 0.1]}\\\cline{2-6}

\multirow{2}{*}{A-GEM}&\multicolumn{5}{c}{LR: [0.0001, 0.0003, 0.001, 0.003, 0.01, 0.03, 0.1]}\\
&\multicolumn{5}{c}{WD: [0.0001, 0.001, 0.01, 0.1]}\\\cline{2-6}
\multirow{3}{*}{CN-DPM}&LR: [0.0001, 0.001, 0.01, 0.1]&[0.001, 0.005, 0.01]& [0.001, 0.01]& [0.001, 0.005, 0.01]&[0.001, 0.01]\\
&$cc$: [0.001, 0.01, 0.1]&[0.001, 0.0015, 0.002]&[0.0005, 0.001, 0.002]&[0.0005, 0.001, 0.002]&[0.0005, 0.001, 0.002]\\
&$\alpha$: [-100, -300, -500]&[-1200, -1000, -800]&[-1200, -1000, -800, -300]&[-15000, -5000, -500]&[-1200, -1000, -800, -300]\\\cline{2-6}
GDumb&\multicolumn{5}{c}{LR: .001, 0.01, 0.1], WD:[0.0001, 0.000001]}\\

\bottomrule
\end{tabular}}
\caption{ Hyperparameter grid for the compared methods.  }
\label{table:hyper}
\end{table*}

\section{Additional Experiments and Results}
\label{additional}
\subsection{More Results for OCI Setting}
\label{additional_oci}
Fig~\ref{fig:cifar_full_mem}, \ref{fig:mini_full_mem} and \ref{fig:core_full_mem} show the 
average accuracy measured by the end of each task on Split CIFAR-100, Mini-ImageNet and CORe50-NC with three different memory buffer sizes (1k, 5k, 10k).

\setcounter{figure}{0}
\begin{figure*}[h]
    \centering
    \includegraphics[
    height=4.5cm]{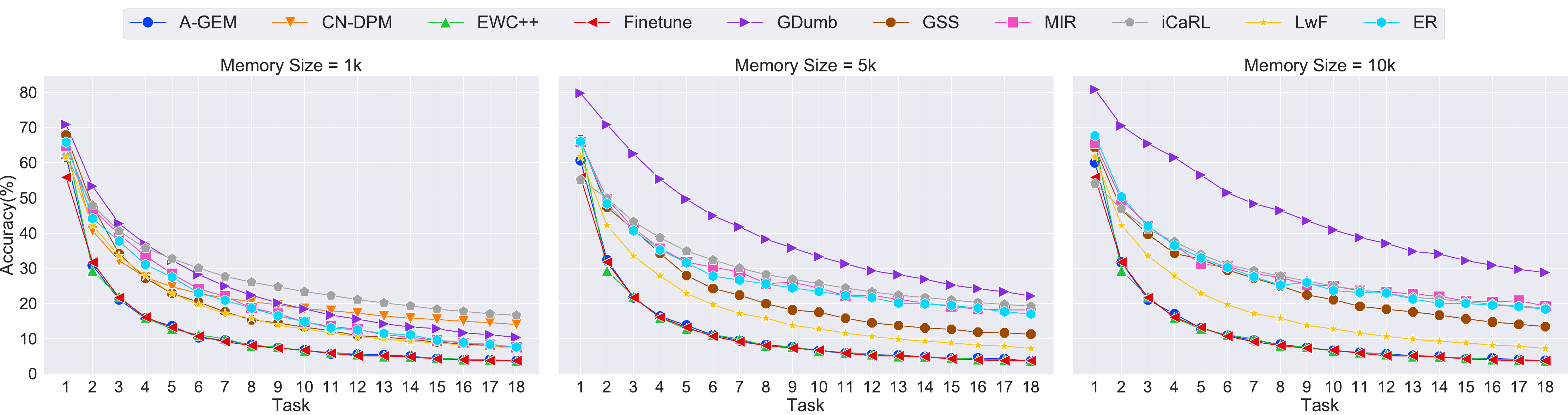}
    \caption{The average accuracy measured by the end of each task for the OCI setting on Split CIFAR-100 with three memory sizes.}\label{fig:cifar_full_mem}
\end{figure*}

\begin{figure*}[h]

    \centering
    \includegraphics[
    height=4.5cm]{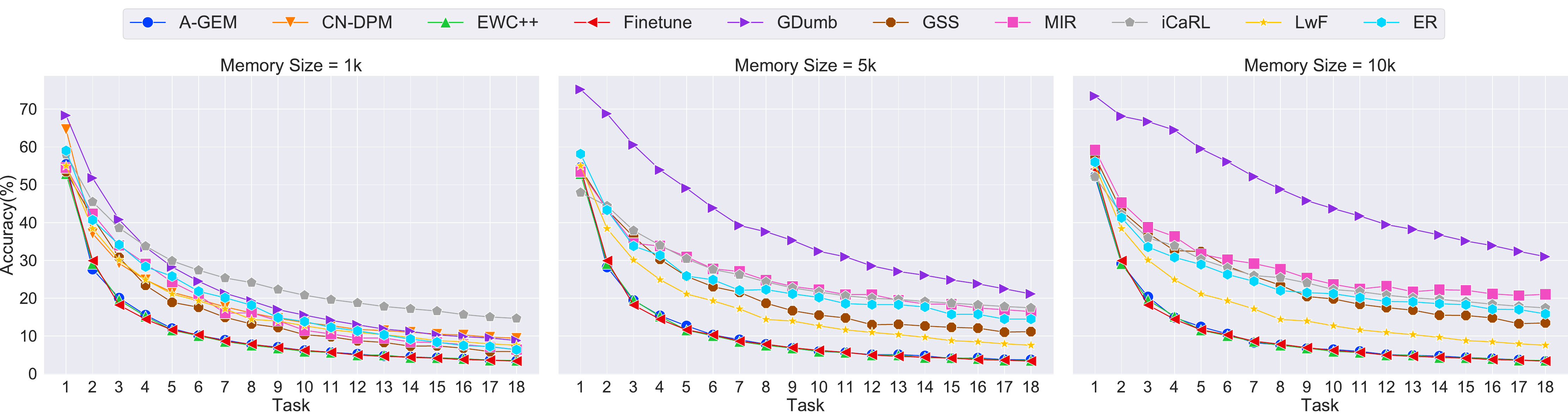}
    \caption{The average accuracy measured by the end of each task for the OCI setting on Split Mini-ImageNet with three memory sizes.}\label{fig:mini_full_mem}
\end{figure*}

\begin{figure*}[h]

    \centering
    \includegraphics[
    height=4.5cm]{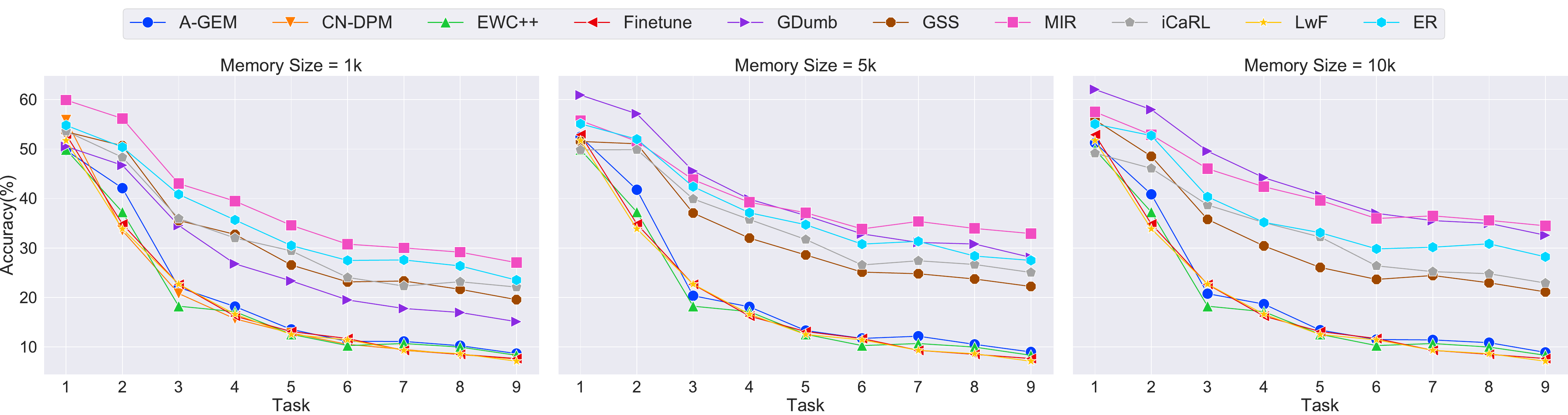}
    \caption{The average accuracy measured by the end of each task for the OCI setting on CORe50-NC with three memory sizes.}\label{fig:core_full_mem}
\end{figure*}

\subsection{OCI Tricks on Split Mini-ImageNet}
\begin{figure*}
    \centering
    \includegraphics[
    height=4.8cm]{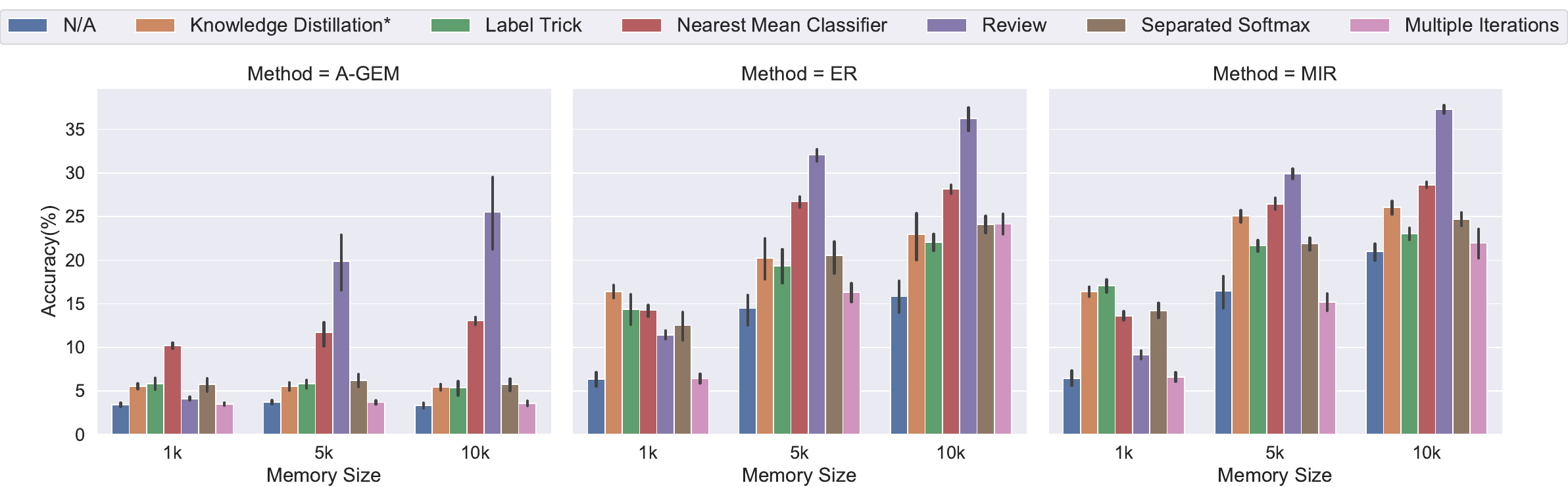}
    \caption{Comparison of various tricks for the OCI setting on Split Mini-ImageNet. We report average accuracy (end of training) for memory buffer with size 1k, 5k and 10k.}\label{fig:trick_mini}
\end{figure*}

We evaluate the tricks described in Section~\ref{trick_detail} on Split Mini-ImageNet. As shown in Table~\ref{table:trick_mini} and Fig.~\ref{fig:trick_mini}, we find similar results as in Split CIFAR-100 that all tricks are beneficial. LB and KDC* are most useful when the memory buffer is small, and NCM and RV are more effective when the memory buffer is large. One main difference is that NCM is not as effective as in CIFAR-100 with a 10k memory buffer as base methods with NCM cannot outperform the best OCI performance.   
\setcounter{table}{0}
\begin{table*}[t!]
\scriptsize
\resizebox{\textwidth}{!}{
\begin{tabular}{l|ccc|ccc|ccc}
\toprule
Finetune & \multicolumn{9}{c}{$3.4\pm0.2$}\\
Offline & \multicolumn{9}{c}{$51.9\pm0.5$}\\
\midrule
Method& \multicolumn{3}{c|}{A-GEM}&\multicolumn{3}{c|}{ER}&\multicolumn{3}{c}{MIR}\\
\midrule
Buffer Size &M=1k& M=5k&M=10k&M=1k& M=5k&M=10k&M=1k&M=5k&M=10k\\
\midrule
    NA&    $3.4\pm0.2$ & $3.7\pm0.3$ & $3.3\pm0.3$& $6.4\pm0.9$ & $14.5\pm2.1$ & $15.9\pm2.0$&$6.4\pm0.9$ & $16.5\pm2.1$ & $21.0\pm1.1$\\
    LB&    $5.8\pm0.8$ & $5.8\pm0.5$ & $5.4\pm0.9$& $14.4\pm2.1$ & $19.3\pm2.3$ & $22.1\pm1.1$&$\mathbf{17.1\pm0.9}$ & $21.7\pm0.7$ & $23.0\pm0.8$\\
    KDC&    $8.0\pm1.1$ & $7.5\pm1.5$ & $8.2\pm1.7$& $12.3\pm2.5$ & $15.4\pm0.4$ & $14.6\pm2.1$&$14.3\pm0.5$ & $15.8\pm0.4$ & $15.5\pm0.5$\\
    KDC*&    $5.6\pm0.4$&$5.5\pm0.5$&$5.4\pm0.4$& $\mathbf{16.4\pm0.8}$ & $20.3\pm2.5$ & $23.0\pm3.1$&$16.4\pm0.6$ & $25.1\pm0.8$ & $26.1\pm0.9$\\
    MI&$3.5\pm0.2$&$3.7\pm0.2$&$3.6\pm0.3$&$6.4\pm0.6$&$16.3\pm1.3$&$24.1\pm1.3$&$6.6\pm0.6$&$15.2\pm1.1$&$22.0\pm1.9$\\
    SS&    $5.7\pm0.9$ & $6.2\pm0.8$ & $5.7\pm0.8$& $12.5\pm1.9$ & $20.5\pm2.1$ & $24.1\pm1.1$&$14.2\pm1.0$ & $21.9\pm0.8$ & $24.7\pm0.9$\\
    RV&$4.1\pm0.2$&$\mathbf{19.9\pm3.7}$&$\mathbf{25.5\pm4.7}$&$11.4\pm0.6$&$\mathbf{32.1\pm0.8}$&$\mathbf{36.3\pm1.5}$&$9.1\pm0.5$&$\mathbf{29.9\pm0.7}$&$\mathbf{37.3\pm0.5}$\\
    NCM&    $\mathbf{10.2\pm0.4}$ & $11.7\pm1.5$ & $13.0\pm0.5$& $14.2\pm0.7$ & $26.7\pm0.7$ & $28.2\pm0.6$&$13.6\pm0.6$ & $26.4\pm0.7$ & $28.6\pm0.4$\\

\midrule
Best OCI & $14.7\pm0.4$&${21.1\pm1.7}$&${31.0\pm1.4}$& $14.7\pm0.4$&${21.1\pm1.7}$&${31.0\pm1.4}$& $14.7\pm0.4$&${21.1\pm1.7}$&${31.0\pm1.4}$\\
\bottomrule
\end{tabular}}
\caption{Performance of compared tricks for the OCI setting on Split Mini-ImageNet. We report average accuracy (end of training) for memory buffer with size 1k, 5k and 10k. Best OCI refers to the best performance from the compared methods in Table~\ref{table:nc}.
}
\label{table:trick_mini}
\end{table*}

\subsection{More Results for ODI Setting}

\begin{figure*}[h]
    \centering
    \includegraphics[
    height=4.5cm]{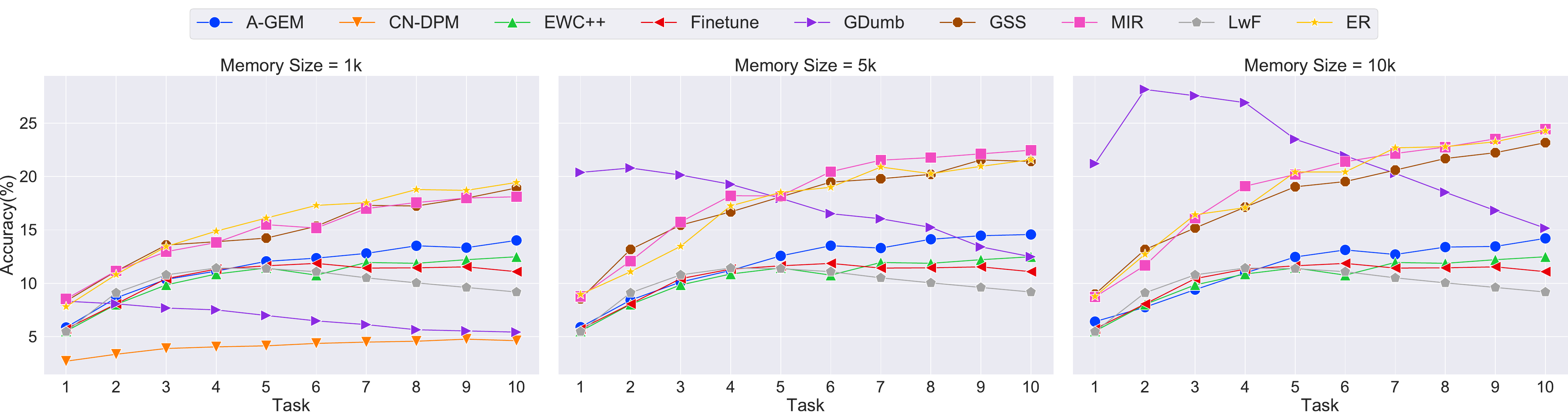}
    \caption{The average accuracy measured by the end of each task for the ODI setting on Mini-ImageNet-Noise with three memory sizes.}\label{fig:noise_allmem}
\end{figure*}

\begin{figure*}[h]

    \centering
    \includegraphics[
    height=4.5cm]{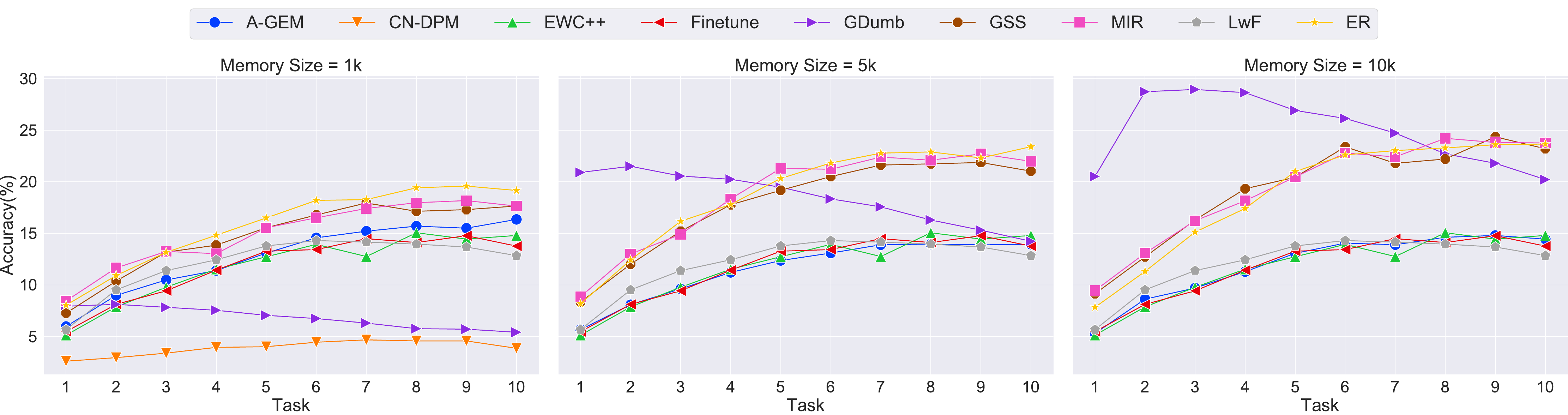}
    \caption{The average accuracy measured by the end of each task for the ODI setting on Mini-ImageNet-Occlusion with three memory sizes.}\label{fig:occlusion_allmem}
\end{figure*}

\begin{figure*}[h]

    \centering
    \includegraphics[
    height=4.5cm]{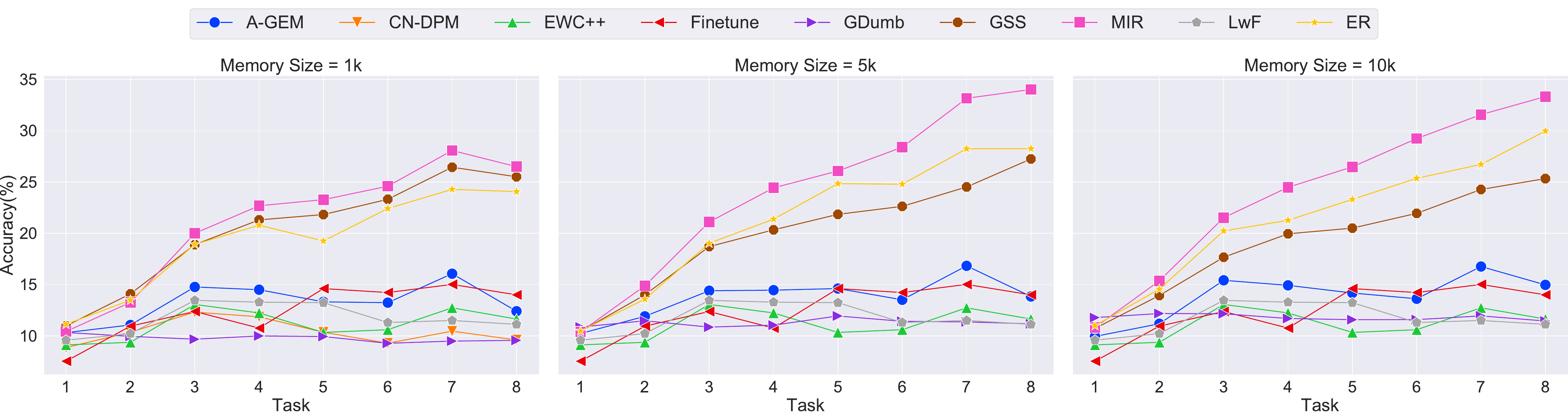}
    \caption{The average accuracy measured by the end of each task for the ODI setting on CORe50-NI with three memory sizes.}\label{fig:coreni_allmem}
\end{figure*}

Fig.~\ref{fig:noise_allmem}, \ref{fig:occlusion_allmem} and \ref{fig:coreni_allmem} show the 
average accuracy measured by the end of each task on Mini-ImageNet-Noise, Mini-ImageNet-Occlusion and CORe50-NI with three different memory buffer sizes (1k, 5k, 10k).

\end{document}